\documentclass[10pt,twocolumn,letterpaper]{article}

\usepackage{cvpr}

\usepackage[dvipsnames]{xcolor}

\definecolor{cvprblue}{rgb}{0.21,0.49,0.74}
\usepackage[pagebackref,breaklinks,colorlinks,citecolor=cvprblue]{hyperref}
\usepackage{microtype}
\usepackage{graphicx}
\usepackage{booktabs}
\usepackage{amssymb}
\usepackage{algorithm,algorithmic}
\usepackage{amsmath}
\usepackage{tabularx}
\usepackage{multirow}
\usepackage{color}
\usepackage{comment}
\usepackage{listings}
\usepackage{multicol} 

\def\paperID{7141} 
\def\confName{CVPR}
\def\confYear{2024}

\title{Towards Understanding Cross and Self-Attention in Stable Diffusion for Text-Guided Image Editing}

\author{Bingyan Liu$^{1,2}$, Chengyu Wang$^2$\thanks{Co-corresponding authors.}, Tingfeng Cao$^{1,2}$, Kui Jia$^3$\footnotemark[1], Jun Huang$^2$\\
$^1$South China University of Technology, Guangzhou, China\\
$^2$ Alibaba Group, Hangzhou, China\\
$^3$The Chinese University of Hong Kong, Shenzhen, China\\
{
\tt\small \{eeliubingyan, setingfengcao\}@mail.scut.edu.cn, 
    }
    {
\tt\small \{chengyu.wcy,huangjun.hj\}@alibaba-inc.com
    }\\
    {
\tt\small kuijia@gmail.com
    }
}

\begin{document}
\maketitle
\begin{abstract}
Deep Text-to-Image Synthesis (TIS) models such as Stable Diffusion have recently gained significant popularity for creative text-to-image generation. However, for domain-specific scenarios, tuning-free Text-guided Image Editing (TIE) is of greater importance for application developers. This approach modifies objects or object properties in images by manipulating feature components in attention layers during the generation process.
Nevertheless, little is known about the semantic meanings that these attention layers have learned and which parts of the attention maps contribute to the success of image editing. In this paper, we conduct an in-depth probing analysis and demonstrate that cross-attention maps in Stable Diffusion often contain object attribution information, which can result in editing failures. In contrast, self-attention maps play a crucial role in preserving the geometric and shape details of the source image during the transformation to the target image. Our analysis offers valuable insights into understanding cross and self-attention mechanisms in diffusion models.
Furthermore, based on our findings, we propose a simplified, yet more stable and efficient, tuning-free procedure that modifies only the self-attention maps of specified attention layers during the denoising process. Experimental results show that our simplified method consistently surpasses the performance of popular approaches on multiple datasets.
\footnote{Source code and datasets are available at \url{https://github.com/alibaba/EasyNLP/tree/master/diffusion/FreePromptEditing}.}
\end{abstract}

\section{Introduction}
\label{sec:intro}
\begin{figure}[h]
\centering
\includegraphics[width=0.465\textwidth]{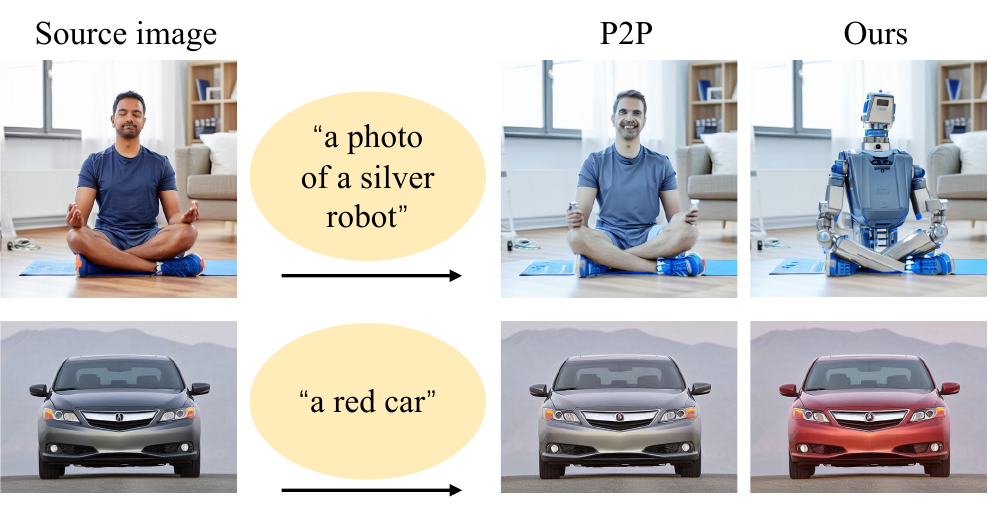}
\caption{An example showing that our method can perform more consistent and realistic TIE compared to P2P~\cite{p2p}.}
\label{fig:intro_shown}
\end{figure}
Text-to-Image Synthesis (TIS) models, such as Stable Diffusion~\cite{LDM}, DALL-E 2~\cite{DALLE2}, and Imagen~\cite{Imagen}, have demonstrated remarkable visual effects for text-to-image generation, capturing substantial attention from both academia and industry~\cite{diffusion_survey1,diffusion_survey2,DALL-E3,DBLP:journals/corr/abs-2309-05534}.
These TIS models are trained on vast amounts of image-text pairs, such as Laion~\cite{Laion_5b,laion_400m}, and employ cutting-edge techniques, including large-scale pre-trained language models~\cite{clip,T5}, variational auto-encoders~\cite{VAE}, and diffusion models~\cite{DDPM,org_diffusion} to achieve success in generating realistic images with vivid details. Specifically, Stable Diffusion stands out as a popular and extensively studied model, making significant contributions to the open-source community.

In addition to image generation, these TIS models possess powerful image editing capabilities, which hold great importance as they aim to modify images while ensuring realism, naturalness, and meeting human preferences.
Text-guided Image Editing (TIE) involves modifying an input image based on a descriptive prompt. Existing TIE methods~\cite{p2p,pnp,null-text-inversion,masactrl,diffedit,instructpix2pix,SDEdit,Shape-guided,pix2pixzero,masactrl} achieve remarkable effects in image translation, style transfer, and appearance replacement, as well as preserving the input structure and scene layout.
To this end, Prompt-to-Prompt (P2P)~\cite{p2p} modifies image regions by replacing cross-attention maps corresponding to the target edit words in the source prompt.
Plug-and-Play (PnP)~\cite{pnp} first extracts the spatial features and self-attention of the original image in the attention layers and then injects them into the target image generation process.
Among these methods, attention layers play a crucial role in controlling the image layout and the relationship between the generated image and the input prompt.
However, inappropriate modifications to attention layers can yield varied editing outcomes and even lead to editing failures. For example, as depicted in Figure~\ref{fig:intro_shown}, editing authentic images on cross-attention layers can result in editing failures; converting a man into a robot or changing the color of a car to red fails.
Moreover, some operations in the above-mentioned methods can be revised and optimized.

In our paper, we explore attention map modification to gain comprehensive insights into the underlying mechanisms of TIE using diffusion-based models. Specifically, we focus on the attribution of TIE and ask the fundamental question:~\emph{how does the modification of attention layers contribute to diffusion-based TIE?}
To answer this question, we carefully construct new datasets and meticulously investigate the impact of modifying the attention maps on the resulting images. This is accomplished by probe analysis~\cite{BERT_probe,linguistic} and systematic exploration of attention map modification with different blocks in the diffusion model.
We find that (1) editing cross-attention maps in diffusion models is optional for image editing. Replacing or refining cross-attention maps between the source and target image generation process is dispensable and can result in failed image editing.
(2) The cross-attention map is not only a weight measure of the conditional prompt at the corresponding positions in the generated image but also contains the semantic features of the conditional token.
Therefore, replacing the target image's cross-attention map with the source image's map may yield unexpected outcomes.
(3) Self-attention maps are crucial to the success of the TIE task, as they reflect the association between image features and retain the spatial information of the image.
Based on our findings, we propose a simplified and effective algorithm called Free-Prompt-Editing (FPE). FPE performs image editing by replacing the self-attention map in specific attention layers during denoising, without needing a source prompt. It is beneficial for real image editing scenarios.
The contributions of our paper are as follows:
\begin{itemize}
\item We conduct a comprehensive analysis of how attention layers impact image editing results in diffusion models and answer why TIE methods based on cross-attention map replacement can lead to unstable results.
\item We design experiments to prove that cross-attention maps not only serve as the weight of the corresponding token on the corresponding pixel but also contain the characteristic information of the token. In contrast, self-attention is crucial in ensuring that the edited image retains the original image's layout information and shape details.
\item Based on our experimental findings, we simplify currently popular tuning-free image editing methods and propose FPE, making the image editing process simpler and more effective.
In experimental tests over multiple datasets, FPE outperforms current popular methods.
\end{itemize}
Overall, our paper contributes to the understanding of attention maps in Stable Diffusion and provides a practical solution for overcoming the limitations of inaccurate TIE.
\section{Related Works}
Text-guided Image Editing (TIE)~\cite{image_editing_survey} is a crucial task 
involving the modification of an input image with requirements expressed by texts.
These approaches can be broadly categorized into two groups: tuning-free methods and fine-tuning based methods.

\subsection{Tuning-free Methods}
Tuning-free TIE methods aim to control the generated image in the denoising process. To achieve this goal, SDEdit~\cite{SDEdit} uses the given guidance image as the initial noise in the denoising step, which leads to impressive results.
Other methods operate in the feature space of diffusion models to achieve successful editing results. One notable example is P2P~\cite{p2p}, which discovers that manipulating cross-attention layers allows for controlling the relationship between the spatial layout of the image and each word in the text. Null-text inversion~\cite{null-text-inversion} further employs an optimization method to reconstruct the guidance image and utilizes P2P for real image editing.
DiffEdit~\cite{diffedit} automatically generates a mask by comparing different text prompts to help guide the areas of the image that need editing.
PnP~\cite{pnp} focuses on spatial features and self-affinities to control the generated image's structure without restricting interaction with the text. Additionally, MasaCtrl~\cite{masactrl} converts self-attention in diffusion models into a mutual and mask-guided self-attention strategy, enabling pose transformation.
In this paper, we aim to provide in-depth insights into the attention layers of diffusion models and further propose a more streamlined tuning-free TIE approach.

\subsection{Fine-tuning Based Methods}
The core idea of fine-tuning-based TIE methods is to synthesize ideal new images by model fine-tuning over the knowledge of domain-specific data~\cite{Dreambooth,textual-inversion,imagic,lora} or by introducing additional guidance information~\cite{controlnet,T2I-adapter,instructpix2pix}.
DreamBooth~\cite{Dreambooth} fine-tunes all the parameters in the diffusion model while keeping the text transformer frozen and utilizes generated images as the regularization dataset.
Textual Inversion~\cite{textual-inversion} optimizes a new word embedding token for each concept.
Imagic~\cite{imagic} learns the approximate text embedding of the input image through tuning and then edits the posture of the object in the image by interpolating the approximate text embedding and the target text embedding.
ControlNet~\cite{controlnet} and T2I-Adapter~\cite{T2I-adapter} allow users to guide the generated images through input images by tuning additional network modules.
Instructpix2pix~\cite{instructpix2pix} fully fine-tunes the diffusion model by constructing image-text-image triples in the form of instructions, enabling users to edit authentic images using instruction prompts, such as ``turn a man into a cyborg''.
In contrast to these works, our method focuses on tuning-free techniques without the fine-tuning process.

\begin{figure}[h]
\centering
\includegraphics[width=0.475\textwidth]{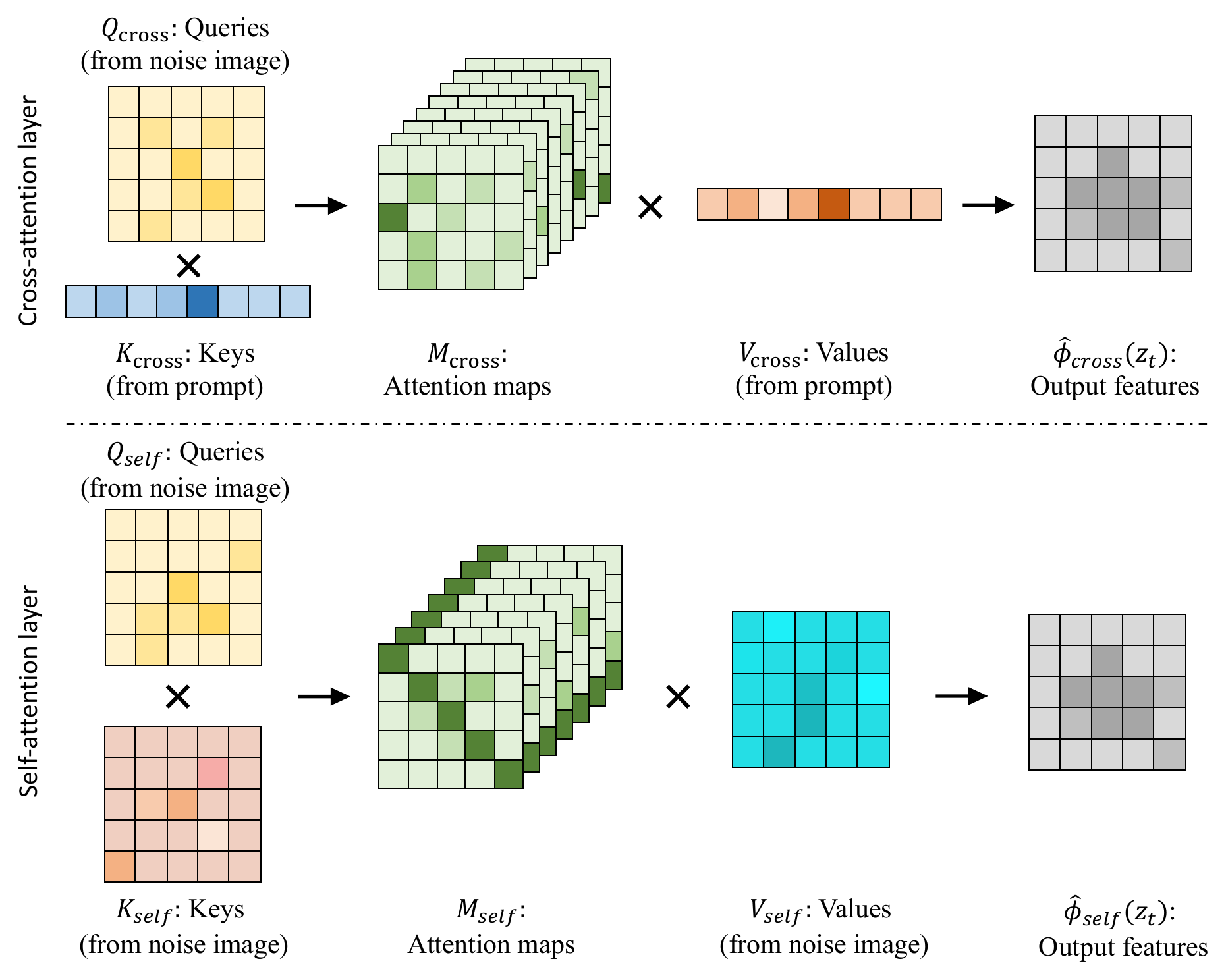}
\caption{Cross and self-attention layers in Stable Diffusion.}
\label{fig:cross_attn}
\end{figure}
\section{Analysis on Cross and Self-Attention}
In this section, we analyze how cross and self-attention maps in Stable Diffusion contribute to the effectiveness of TIE.

\subsection{Cross-Attention in Stable Diffusion}
In Stable Diffusion and other similar models, cross-attention layers play a crucial role in fusing images and texts, allowing T2I models to generate images that are consistent with textual descriptions. As depicted in the upper part of Figure~\ref{fig:cross_attn}, the cross-attention layer receives the query, key, and value matrices, i.e., $Q_{cross}$, $K_{cross}$, and $V_{cross}$, from the noisy image and prompt. Specifically, $Q_{cross}$ is derived from the spatial features of the noisy image $\phi_{cross}(z_{t})$ by learned linear projections $\ell_{q}$, while $K_{cross}$ and $V_{cross}$ are projected from the textual embedding $P_{emb}$ of the input prompt $P$ using learned linear projections denoted as $\ell_{k}$ and $\ell_{v}$, respectively.
The cross-attention map is defined as:
\begin{equation}
 Q_{cross} = \ell_{q}(\phi_{cross}(z_{t})), \quad K_{cross} = \ell_{k}(P_{emb})
\label{equation3}
\end{equation}
\begin{equation}
 M_{cross} = \text{Softmax}\left(\frac{Q_{cross}{K_{cross}}^{\mathsf{T}}}{\sqrt{d_{cross}}}\right)
\label{equation4}
\end{equation}
where $d_{cross}$ is the dimension of the keys and queries.
The final output is defined as the fused feature of the text and image, denoted as $\hat{\phi}(z_{t}) = M_{cross}V_{cross}$, where $V_{cross} = \ell_{v}(P_{emb})$.
Intuitively, each cell in the cross-attention map, denoted as $M_{ij}$, determines the weights attributed to the value of the $j$-th token relative to the spatial feature $i$ of the image. The cross-attention map enables the diffusion model to 
locate/align the tokens of the prompt in the image area.

\subsection{Self-Attention in Stable Diffusion}
As depicted in Figure~\ref{fig:cross_attn}, unlike cross-attention, the self-attention layer receives the keys matrix $K_{self}$ and the query matrix $Q_{self}$ from the noisy image $\phi_{self}(z_{t})$ through learned linear projections $\bar\ell_{K}$ and $\bar\ell_{Q}$, respectively.
The self-attention map is defined as:
\begin{equation}
 Q_{self} = \bar\ell_{q}(\phi_{self}(z_{t})), \quad K_{self} = \bar\ell_{K}(\phi_{self}(z_{t})),
\label{equation5}
\end{equation}
\begin{equation}
 M_{self} = \text{Softmax}\left(\frac{Q_{self}{K_{self}}^{\mathsf{T}}}{\sqrt{d_{self}}}\right)
\label{equation6}
\end{equation}
where $d_{self}$ is the dimension of $K_{self}$ and $Q_{self}$.
$M_{self}$ determines the weights assigned to the relevance of the $i$-th and $j$-th spatial features in the image and can affect the spatial layout and shape details of the generated image. Consequently, the self-attention map can be utilized to preserve the spatial structure characteristics of the original image throughout the image editing process.

\subsection{Probing Analysis}
Yet, the semantics of cross and self-attention maps remain unclear. 
Are these attention maps merely weight matrices, or do they contain feature information of the image? To answer these questions, we aim to explore the meaning of attention maps in diffusion models. Inspired by probing analysis methods~\cite{BERT_probe,linguistic} in the field of NLP, we propose building datasets and training classification networks to explore the properties of attention maps.
Our fundamental idea is that if a trained classifier can accurately classify attention maps from different categories, then the attention map contains meaningful feature representation of the category information. Therefore, we introduce a task-specific classifier on top of the diffusion model's cross-attention and self-attention layers. This classifier is a two-layer MLP designed to predict specific semantic properties of the attention maps.
To present the analysis results more visually, we utilize color adjectives and animal nouns to form prompt datasets, each containing ten categories. For the color adjective, there are two prompt formats: \emph{a $<$color$>$ car} and \emph{a $<$color$>$ $<$object$>$}. The prompt format for animal nouns is \emph{a/an $<$animal$>$ standing in the park}. After generating the prompts, we employ the probing method to extract the cross-attention maps corresponding to the words $<$color$>$ and $<$animal$>$, along with the self-attention maps in the attention layers.
Finally, by training and evaluating the performance of the classifiers, we gain insights into the semantic knowledge captured by the attention maps.

\begin{figure}[ht]
\centering
\includegraphics[width=0.485\textwidth]{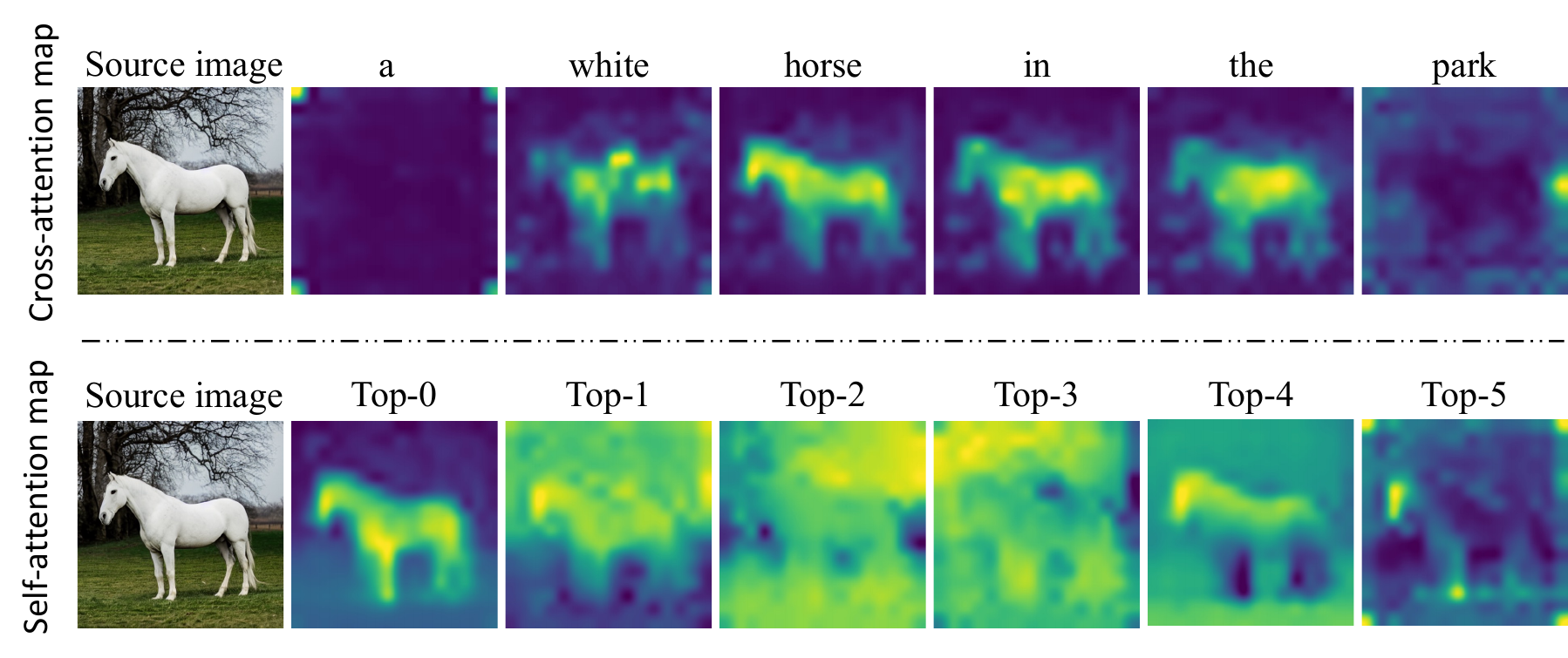}
\caption{The heatmaps of cross-attention and self-attention maps in a generated image with the prompt "a white horse in the park". The visualization of the cross-attention map corresponds to each word in the prompt. The visualization of the self-attention map is the top-6 components obtained after SVD~\cite{SVD}.}
\label{fig:attn_map}
\end{figure}
\subsection{Probing Results on Cross-Attention Maps}
\label{Probing_Cross-Attention}
\begin{figure}[h]
\centering
\includegraphics[width=0.495\textwidth]{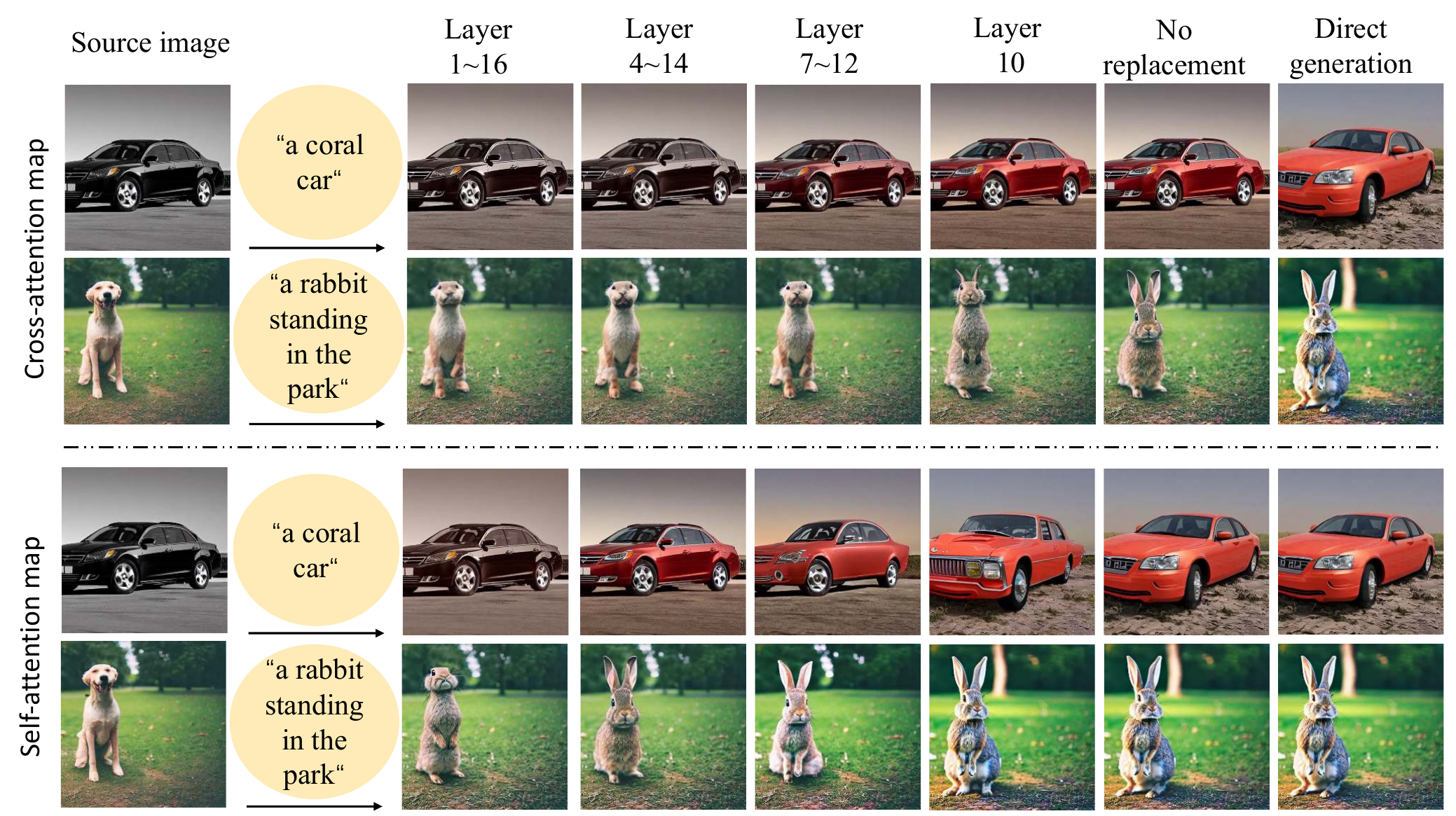}
\caption{Results of cross-attention and self-attention map replacements in difference layers of the diffusion model.}
\label{fig:case_cross_and_self_attn}
\end{figure}

\noindent\textbf{What does the cross-attention map learn?} 
We directly visualize the attention maps, as demonstrated in Figure~\ref{fig:attn_map}. Each word in the prompt has a corresponding attention map associated with the image, indicating that the information related to the word exists in specific areas of the image. However, is this information exclusive to these areas? Referring to Equation~\ref{equation4}, we observe that $M_{cross}$ is derived from $K_{cross}$ and $Q_{cross}$, indicating that $M_{cross}$ carries information from both. 
To validate this hypothesis, we conduct probing experiments on $M_{cross}$, with the results presented in Table~\ref{cross_attn_result}. Due to space limitations, we show only the probing results for five colors and five animals from the last layer of the down, middle, and up blocks. 
As evident in Table~\ref{cross_attn_result}, the trained classifier achieves high accuracy in both the color and animal classification tasks. For instance, the average accuracy for classifying ``sheep'' reaches 98\%, and that for ``orange'' reaches 93\%.
These results demonstrate that the cross-attention map acts as a reliable category representation, indicating that it reflects not only weight information but also contains category-related features. This explains the failure of image editing using cross-attention map replacement. The upper part of Figure~\ref{fig:case_cross_and_self_attn} illustrates the editing results obtained by replacing the cross-attention map of the corresponding word (``rabbit'' and ``coral'') at different cross-attention layers. It is apparent that when all layers are replaced, the editing results are the least satisfactory. The dog fails to transform completely into a rabbit, and the black car cannot turn into a coral car. Conversely, when the cross-attention map is left unaltered, correct editing results can be achieved. The complete and more additional experimental results are available in Section~\ref{sub:Probing Analysis} in the Supplementary Material.

\begin{table}[h]
\centering
\aboverulesep=0pt
\belowrulesep=0pt
\begin{scriptsize}
\begin{tabularx}{0.495\textwidth}{l@{\hspace{4pt}}c@{\hspace{4pt}}c@{\hspace{4pt}}c@{\hspace{4pt}}c@{\hspace{4pt}}c@{\hspace{4pt}}c@{\hspace{4pt}}c|@{\hspace{4pt}}c}
\toprule
Class  & Layer 3  & Layer 6 & Layer 9 & Layer 10  & Layer 12 & Layer 14 & Layer 16 & Avg. \\
\hline
dog & 1.00 & 1.00 & 1.00 & 1.00 & 0.89 & 0.76 & 1.00 & 0.95 \\
horse & 0.96 & 1.00 & 1.00 & 1.00 & 0.64 & 1.00 & 0.91 & 0.93 \\
sheep & 0.97 & 1.00 & 1.00 & 1.00 & 1.00 & 0.90 & 0.97 & 0.98 \\
leopard & 0.97 & 1.00 & 1.00 & 1.00 & 0.97 & 0.79 & 0.87 & 0.94 \\
tiger & 1.00 & 1.00 & 0.97 & 1.00 & 0.88 & 1.00 & 0.97 & 0.97 \\
\midrule
green & 0.93 & 0.91 & 0.91 & 0.96 & 0.67 & 0.38 & 0.60 & 0.77 \\
white & 0.97 & 1.00 & 0.94 & 0.97 & 0.97 & 0.61 & 0.85 & 0.90 \\
orange & 0.97 & 1.00 & 0.94 & 0.92 & 0.89 & 0.94 & 0.83 & 0.93 \\
yellow & 0.96 & 0.77 & 1.00 & 0.98 & 1.00 & 0.36 & 0.68 & 0.82 \\
red & 0.97 & 0.97 & 0.93 & 0.85 & 0.70 & 0.23 & 0.65 & 0.76 \\
\bottomrule
\end{tabularx}
\end{scriptsize}
\caption{Probing accuracy of cross-attention map in difference layers.}
\label{cross_attn_result}
\end{table}

\begin{table}[h]
\centering
\aboverulesep=0pt
\belowrulesep=0pt
\begin{scriptsize}
\begin{tabularx}{0.495\textwidth}{l@{\hspace{4pt}}c@{\hspace{4pt}}c@{\hspace{4pt}}c@{\hspace{4pt}}c@{\hspace{4pt}}c@{\hspace{4pt}}c@{\hspace{4pt}}c|@{\hspace{4pt}}c}
\toprule
Class  & Layer 3  & Layer 6 & Layer 9 & Layer 10  & Layer 12 & Layer 14 & Layer 16 & Avg. \\
\hline
dog & 0.53 & 0.60 & 0.78 & 0.60 & 0.53 & 0.47 & 0.38 & 0.55 \\
horse & 0.50 & 0.70 & 0.82 & 0.65 & 0.68 & 0.53 & 0.28 & 0.59 \\
sheep & 0.53 & 0.45 & 0.25 & 0.45 & 0.62 & 0.53 & 0.25 & 0.44 \\
leopard & 0.47 & 0.65 & 0.57 & 0.60 & 0.47 & 0.65 & 0.60 & 0.57 \\
tiger & 0.23 & 0.12 & 0.55 & 0.20 & 0.45 & 0.42 & 0.53 & 0.36 \\
\midrule
green & 0.00 & 0.00 & 0.05 & 0.00 & 0.05 & 0.00 & 0.12 & 0.03 \\
white & 0.00 & 0.05 & 0.30 & 0.55 & 0.03 & 0.15 & 0.25 & 0.19 \\
orange & 0.00 & 0.00 & 0.00 & 0.00 & 0.00 & 0.00 & 0.00 & 0.00 \\
yellow & 0.00 & 0.42 & 0.07 & 0.05 & 0.00 & 0.30 & 0.07 & 0.13 \\
red & 0.00 & 0.15 & 0.28 & 0.20 & 0.00 & 0.20 & 0.10 & 0.13 \\
\bottomrule
\end{tabularx}
\end{scriptsize}
\caption{Probing accuracy of self-attention map in difference layers.}
\label{self_attn_result}
\end{table}

\subsection{Probing Results on Self-Attention Maps}
\noindent\textbf{What does the self-attention map learn?} Table~\ref{self_attn_result} presents the results of the probing experiments. The results indicate that the trained classifier struggles to classify the self-attention map generated from images containing color prompts.
For animals, the results are better, although not as precise as those using cross-attention maps. 
This discrepancy may be attributed to the irregular spatial structure present in the self-attention map corresponding to the color prompt. Conversely, the self-attention map corresponding to the animal prompt contains structural information of different animals, enabling the learning of category information through recognizing structural or contour features. As shown in the lower part of Figure~\ref{fig:attn_map}, the first component of the horse's self-attention map clearly expresses the outline information of the horse.
The lower part of Figure~\ref{fig:case_cross_and_self_attn} showcases our experimental results of operating on the self-attention map across different attention layers. When the self-attention map of all layers in the source image is replaced during the generation process of the target image, the resulting target image retains all the structural information from the original image but hinders successful editing. Conversely, if we do not replace the self-attention map, we obtain an image identical to that generated directly using the target prompt. As a compromise, replacing the self-attention map in Layers 4 to 14 allows for preserving the structural information of the original image to the greatest extent while ensuring successful editing. 
This experimental result further supports the idea that the self-attention map in Layers 4 to 14 does not serve as a reliable category representation but does contain valuable spatial structure information of the image.

\subsection{Probing Results for Other Tokens}
\begin{figure}[h]
\centering
\includegraphics[width=0.485\textwidth]{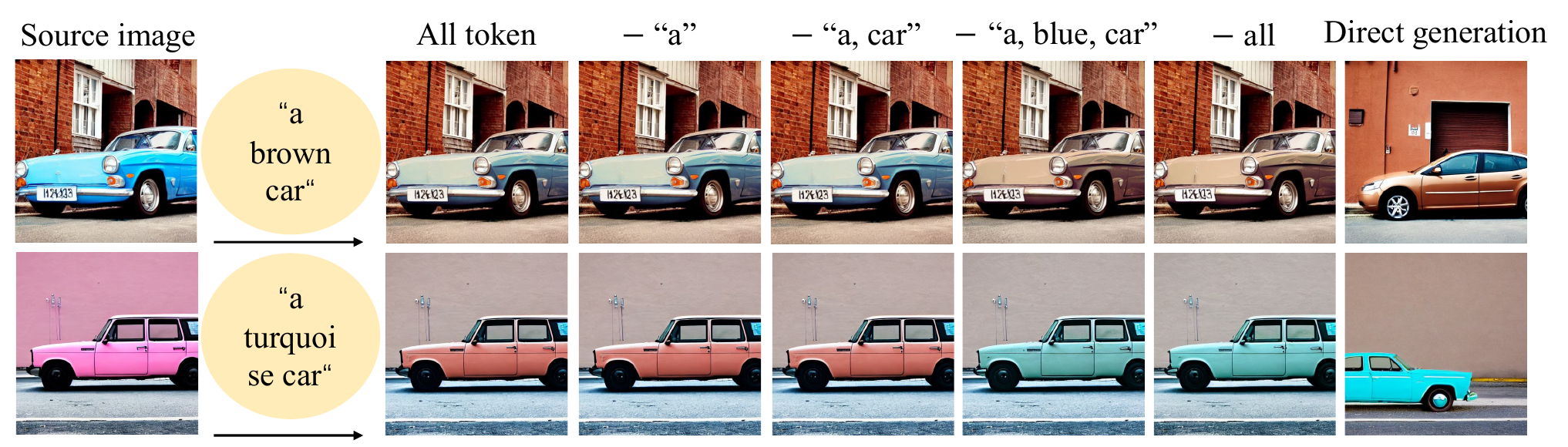}
\caption{Editing results on replacing attention maps of different tokens in a prompt. ``-'' is a minus sign. \emph{- ``a''} represents subtracting the cross-attention map corresponding to ``a''.}
\label{fig:case_diff_token}
\end{figure}

\noindent\textbf{Do cross-attention maps corresponding to non-edited words contain category information?} Furthermore, we explore the attention maps associated with non-edited words. This is relevant because within a text sequence, the text embedding for each word retains the contextual information of the sentence, particularly when a transformer-based text encoder~\cite{clip,bert} is utilized. We employ the prompt data in the format of \emph{a $<$color$>$ car} for our probing experiments. The experimental results are presented in Table~\ref{cross_attn_result_diff_tokens}. The findings demonstrate that the article ``a'' does not encompass any category information of color. In contrast, the noun ``car,'' when modified by the color adjective, does contain color category information. Consequently, if we replace the cross-attention map corresponding to a non-edited word with the cross-attention map of the target image, color information may be introduced, ultimately resulting in editing failures. This observation is also evident from the experimental results in Figure~\ref{fig:case_diff_token}, where replacing the cross-attention maps of non-edited words likewise leads to editing failures.

\begin{table}[htb]
\centering
\aboverulesep=0pt
\belowrulesep=0pt
\begin{scriptsize}
\begin{tabularx}{0.485\textwidth}{l@{\hspace{4pt}}c@{\hspace{4pt}}c@{\hspace{4pt}}c@{\hspace{4pt}}c@{\hspace{4pt}}c@{\hspace{4pt}}c@{\hspace{4pt}}c|@{\hspace{4pt}}c}
\toprule
Class  & Layer 3  & Layer 6 & Layer 9 & Layer 10  & Layer 12 & Layer 14 & Layer 16 &Avg. \\
\hline
green & 0.03 & 0.00 & 0.00 & 0.00 & 0.00 & 0.00 & 0.00 & 0.00 \\
white & 0.20 & 0.00 & 0.70 & 0.00 & 0.72 & 0.30 & 0.05 & 0.28 \\
orange & 0.00 & 0.00 & 0.00 & 0.00 & 0.00 & 0.00 & 0.00 & 0.00 \\
yellow & 0.12 & 0.00 & 0.00 & 0.00 & 0.00 & 0.00 & 0.00 & 0.02 \\
red & 0.50 & 0.00 & 0.82 & 0.00 & 0.00 & 0.00 & 0.00 & 0.19 \\
\midrule
green & 0.67 & 0.67 & 0.00 & 0.00 & 0.00 & 0.50 & 0.00 & 0.26 \\
white & 0.33 & 1.00 & 0.83 & 0.58 & 0.00 & 0.00 & 1.00 & 0.54 \\
orange & 0.60 & 1.00 & 0.80 & 1.00 & 1.00 & 0.40 & 0.80 & 0.80 \\
yellow & 0.50 & 0.25 & 0.00 & 0.00 & 0.12 & 0.25 & 0.00 & 0.16 \\
red & 0.38 & 0.88 & 0.75 & 0.12 & 0.12 & 0.38 & 0.00 & 0.38 \\
\bottomrule
\end{tabularx}
\end{scriptsize}
\caption{Probing analysis of cross-attention maps w.r.t. difference tokens. The upper part shows the classification results corresponding to the token ``a'', and the lower shows results for ``car''.}
\label{cross_attn_result_diff_tokens}
\end{table}

\section{Our Approach}
Based on our exploration of attention layers, we propose a more straightforward yet more stable and efficient approach named Free-Prompt-Editing (FPE).
Let $I_{src}$ be the image to be edited.
Our goal is to synthesize a new desired image $I_{dst}$ based on the target prompt $P_{dst}$ while preserving the content and structure of the original image $I_{src}$. 
Current editing methods like P2P~\cite{p2p} replace the cross-attention map in the source and target image generation process. This requires modifying the original prompt to find the corresponding attention map for replacement. However, this limitation prevents the direct application of P2P to editing real images, as they do not come with an original prompt. 

Based on our exploration of attention layers, our core idea is to combine the layout and contents of $I_{src}$ with the semantic information synthesized with the target prompt $P_{dst}$ to synthesize the desired image $I_{dst}$ that retains the structure and content information of the original image $I_{src}$. To achieve this, we adapt the self-attention hijack mechanism in the diffusion model's attention layers 4 to 14 during the denoising process between the source and target images.
For generated image editing, we substitute the target image's self-attention map with the source image's self-attention map during the diffusion denoising process. When working with actual images, we first obtain the necessary latents for reconstructing the real image by employing the inversion operation~\cite{DDIM}. Subsequently, during the editing process, we replace the self-attention map of the real image within the generation process of the target image.
We can accomplish the TIE task for the following reasons: 1) the cross-attention mechanism~\cite{LDM,attention} facilitates the fusion of the synthetic image and the target prompt, allowing the target prompt and the image to be automatically aligned even without introducing the cross-attention map of the source prompt; 
2) the self-attention map contains spatial layout and shape details of the source image, and the self-attention mechanism~\cite{attention} allows for the injection of structural information from the original image into the generated target image.
Algorithms~\ref{Gen_algorithm} and \ref{Real_algorithm} present the pseudocode for our simplified method applied to generated and real images, respectively. FPE can also be combined with null text inversion for real image editing (refer to Section~\ref{sup:null_text_inversion}).

\begin{algorithm}[!ht]
\small
    \renewcommand{\algorithmicrequire}{\textbf{Input:}}
	\renewcommand{\algorithmicensure}{\textbf{Output:}}
	\caption{Free-Prompt-Editing for a generated image.}
    \label{Gen_algorithm}
    \begin{algorithmic}[1] 
        \REQUIRE  $P_{src}$: a source prompt;
                  $P_{dst}$: a target prompt;
                  $S$: random seed;
	    \ENSURE $I_{src}$: source image;
                $I_{dst}$: edited image;
        
        \STATE $z_{T} \sim \mathcal{N}(0,1)$, a unit Gaussian random value sampled with random seed $S$;
        \STATE $z^{*}_{T} \leftarrow z_{T}$;
        
        \FOR {$t = T, T-1, \ldots, 1$}
            \STATE $z_{t-1}, M_{self} \leftarrow DM(z_{t}, P_{src}, t)$;
            \STATE $z^{*}_{t-1} \leftarrow DM(z^{*}_{t}, P_{dst}, t)\{M^{*}_{self} \leftarrow M_{self}\}$;
        \ENDFOR
        \STATE \textbf{Return} $(I_{src} \leftarrow Generate(z_{0}), I_{dst} \leftarrow Generate(z^{*}_{0}))$;
    \end{algorithmic}
\end{algorithm}

\begin{algorithm}[!ht]
\small
    \renewcommand{\algorithmicrequire}{\textbf{Input:}}
    \renewcommand{\algorithmicensure}{\textbf{Output:}}
    \caption{Free-Prompt-Editing for a real image.}
    \label{Real_algorithm}
    \begin{algorithmic}[1] 
        \REQUIRE  $P_{dst}$: a target prompt;
                  $I_{src}$: real image;
        \ENSURE   $I_{dst}$: edited image;
                  $I_{res}$: reconstructed image;
        
        \STATE $\{z_{t}\}_{t=0}^{T} \leftarrow DDIM-inv(I_{src})$; 
        \STATE $z^{*}_{T} \leftarrow z_{T}$; 
        
        \FOR {$t = T, T-1, \ldots, 1$}
            \STATE $z_{t-1}, M_{self} \leftarrow DM(z_{t}, t)$; 
            \STATE $z^{*}_{t-1} \leftarrow DM(z^{*}_{t}, P_{dst}, t)\{M^{*}_{self} \leftarrow M_{self}\}$; 
        \ENDFOR
        
        \STATE \textbf{Return} $(I_{res} \leftarrow Generate(z_{0}), I_{dst} \leftarrow Generate(z^{*}_{0}))$; 
    \end{algorithmic}
\end{algorithm}

\section{Experiments}
\subsection{Experimental Settings}
Since there are no publicly available datasets specifically designed to verify the effectiveness of image editing algorithms, we construct two types of image-prompt pairs datasets: one for generated images and one for real images. The generated images dataset includes Car-fake-edit and ImageNet-fake-edit, where Car-fake-edit contains 756 prompt pairs, and ImageNet-fake-edit contains 1182 prompt pairs sampled from FlexIT~\cite{LPIPS} and ImageNet~\cite{imagenet_date}. 
The real image datasets include Car-real-edit, sampled from the Stanford Car (CARS196) dataset~\cite{stanfordcar}, containing 3321 image-prompt pairs, and ImageNet-real-edit, which contains 1092 pairs.
For more details, see section~\ref{sub:data_collection_editing_experiments} in the Supplementary Material. 
In addition, we also use benchmarks constructed by PnP~\cite{pnp}. These benchmarks contain two datasets: Wild-TI2I and ImageNet-R-TI2I. For generated images, Wild-TI2I contains 70 prompt pairs, and ImageNet-R-TI2I contains 150 pairs. For real images, Wild-TI2I contains 78 image-prompt pairs, and ImageNet-R-TI2I includes 150 pairs.

We utilize Clip Score (CS) and Clip Directional Similarity (CDS)~\cite{clip,StyleGAN-NADA} to quantitatively analyze and compare our method with currently popular image editing algorithms. 
The underlying model for our experiments is Stable Diffusion 1.5\footnote{\url{https://huggingface.co/runwayml/stable-diffusion-v1-5}}. The experimental results of comparative methods are produced using the publicly disclosed codes from their original papers with unified random seeds.

\subsection{Image Editing Results}
\begin{figure}[h]
\centering
\includegraphics[width=0.455\textwidth]{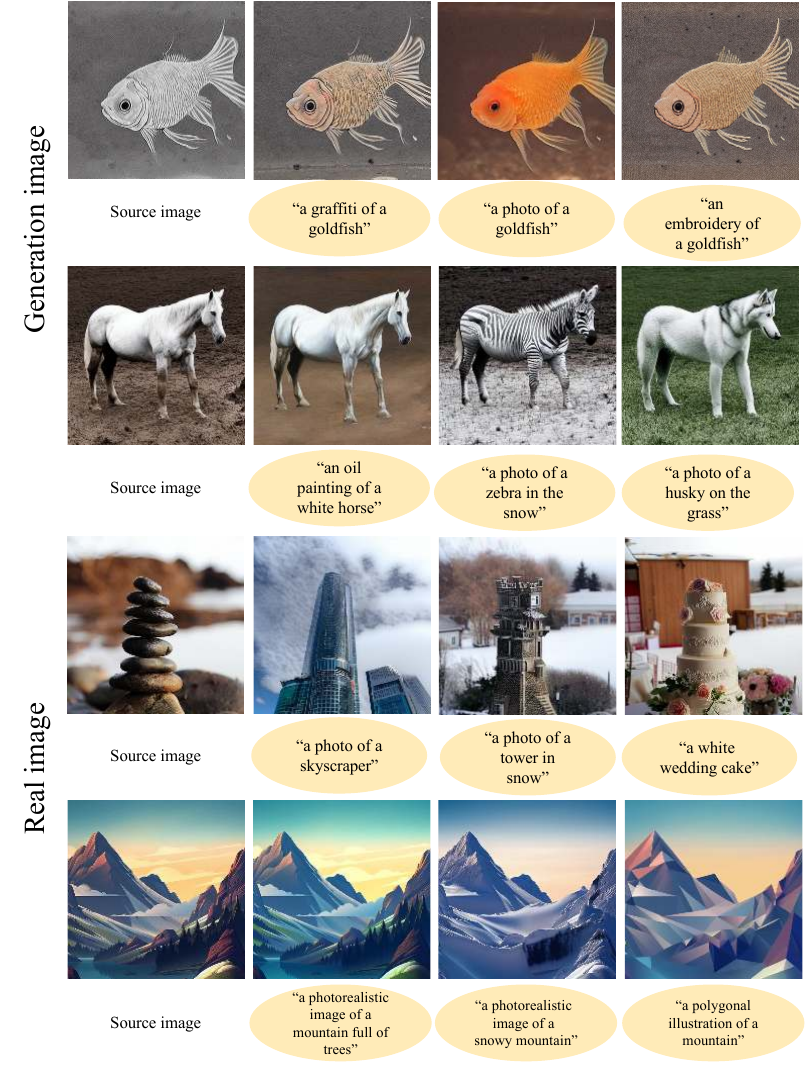}
\caption{Results of our method on image-text pairs from Wild-TI2I and ImageNet-R-TI2I.}
\vspace{-1em}
\label{fig:shown_cases}
\end{figure}

We evaluate our method through quantitative and qualitative analyses. 
As illustrated in Figure~\ref{fig:shown_cases}, we showcase the editing outcomes of our method, demonstrating that it successfully transforms various attributes, styles, scenes, and categories of the original images.

\subsubsection{Comparison to Prior/Concurrent Work}
In this section, we compare our work with state-of-the-art image editing methods, including (i) P2P~\cite{p2p} (with null text inversion~\cite{null-text-inversion} for the real image scene), (ii) PnP~\cite{pnp}, (iii) SDEdit~\cite{SDEdit} under two noise levels (0.5 and 0.75), (iv) DiffEdit~\cite{diffedit}, (v) MasaCtrl~\cite{masactrl}, (vi) Pix2pixzero~\cite{pix2pixzero}, (vii) Shape-guided~\cite{Shape-guided}, and (viii) InstructPix2Pix~\cite{instructpix2pix}. We further present the image editing results using other Stable Diffusion-based models to demonstrate the universality of our method, including Realistic-V2\footnote{\url{https://huggingface.co/SG161222/Realistic_Vision_V2.0}}, Deliberate\footnote{\url{https://huggingface.co/XpucT/Deliberate}}, and Anything-V4\footnote{\url{https://huggingface.co/xyn-ai/anything-v4.0}}.

\begin{figure}[htp]
\centering
\includegraphics[width=0.475\textwidth]{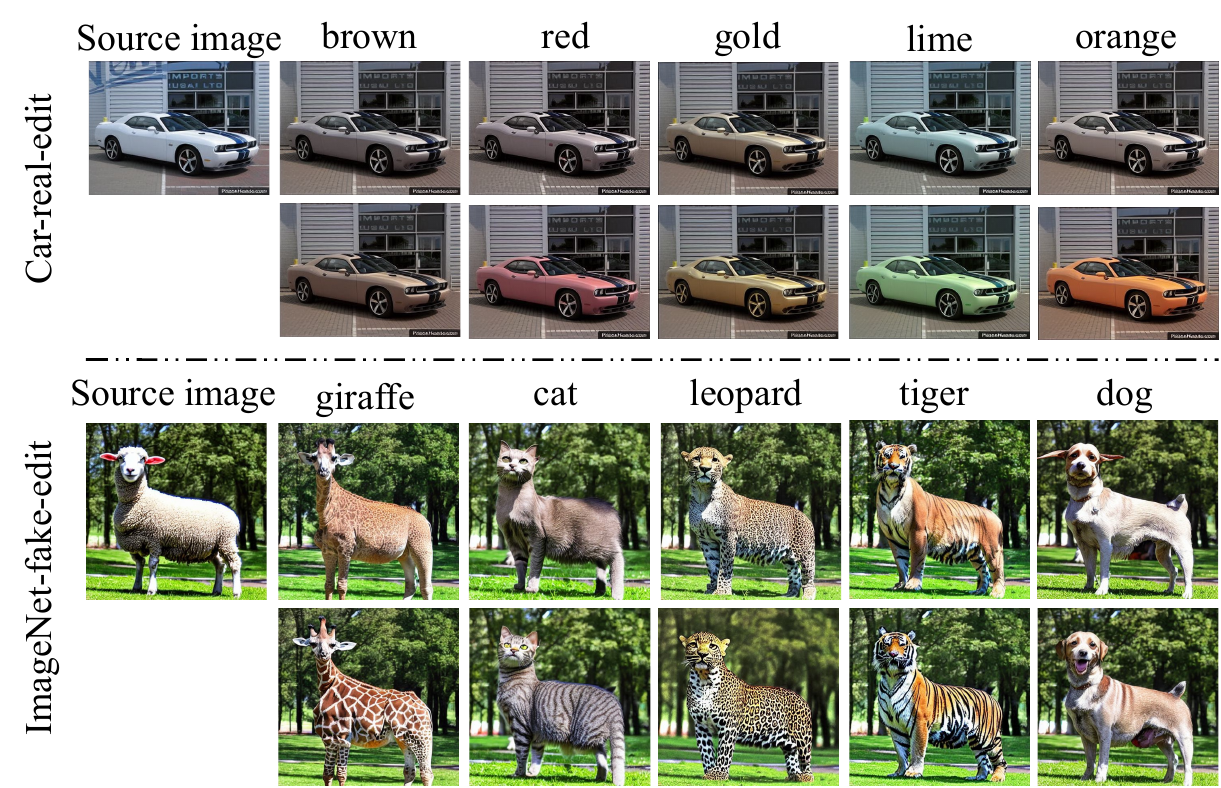}
\caption{Comparisons results with P2P~\cite{p2p} on Car-real-edit and ImageNet-fake-edit. Upper part: P2P. Lower part: ours.}
\label{fig:compaired_w_p2p}
\end{figure}

\noindent\textbf{Comparison to P2P}
We first compare our method with P2P~\cite{p2p} for synthetic image editing scenes and P2P combined with null text inversion~\cite{null-text-inversion} for real image scenes, both denoted as P2P. The experimental results are shown in Figure~\ref{fig:compaired_w_p2p} and Table~\ref{Comparison_results_with_p2p}.
In Figure~\ref{fig:compaired_w_p2p}, it is evident that when performing color transformation on a real image by modifying the cross-attention map, the editing fails. The editing results of P2P for car color tend to replicate the color (white) of the original image. 
Regarding the category conversion results for generated images, we observe that while P2P can accurately transform different animals, the edited results still retain appearances of sheep. This leads to an incomplete conversion for patterned animals such as giraffes, leopards, and tigers.
Unlike P2P, our method operates only at the self-attention layers and is not susceptible to editing failures caused by modifications to the cross-attention map.

\begin{table}[htp]
\centering
\aboverulesep=0pt
\belowrulesep=0pt
\scriptsize
\resizebox{0.465\textwidth}{!}{
\begin{tabular}{l|cc|cc}
\toprule
\multirow{2}{*}{\textbf{Dataset}} & \multicolumn{2}{c|}{\textbf{CS $\uparrow$ }}&\multicolumn{2}{c}{\textbf{CDS $\uparrow$ }}  \\

\cline{2-5}& P2P & Ours & P2P & Ours   \\
\hline
\textbf{Car-fake-edit} &25.96	&\textbf{26.02}   & 0.2451	&\textbf{0.2659}	\\ 

\textbf{Car-real-edit} &24.64	&\textbf{24.85}   &0.2288	&\textbf{0.2605}	\\ 

\textbf{ImageNet-fake-edit} &27.42	& \textbf{27.80}	  &0.2401	&\textbf{0.2560} \\
\textbf{ImageNet-real-edit} &26.17	& \textbf{26.35}	  &0.2426	&\textbf{0.2468}\\ 
\bottomrule
\end{tabular}
}
\caption{Quantitative experimental results over Car-fake-edit, ImageNet-fake-edit, Car-real-edit and ImageNet-real-edit.}
\vspace{-1em}
\label{Comparison_results_with_p2p}
\end{table}

\begin{figure*}[htp]
\centering
\includegraphics[width=0.92\textwidth]{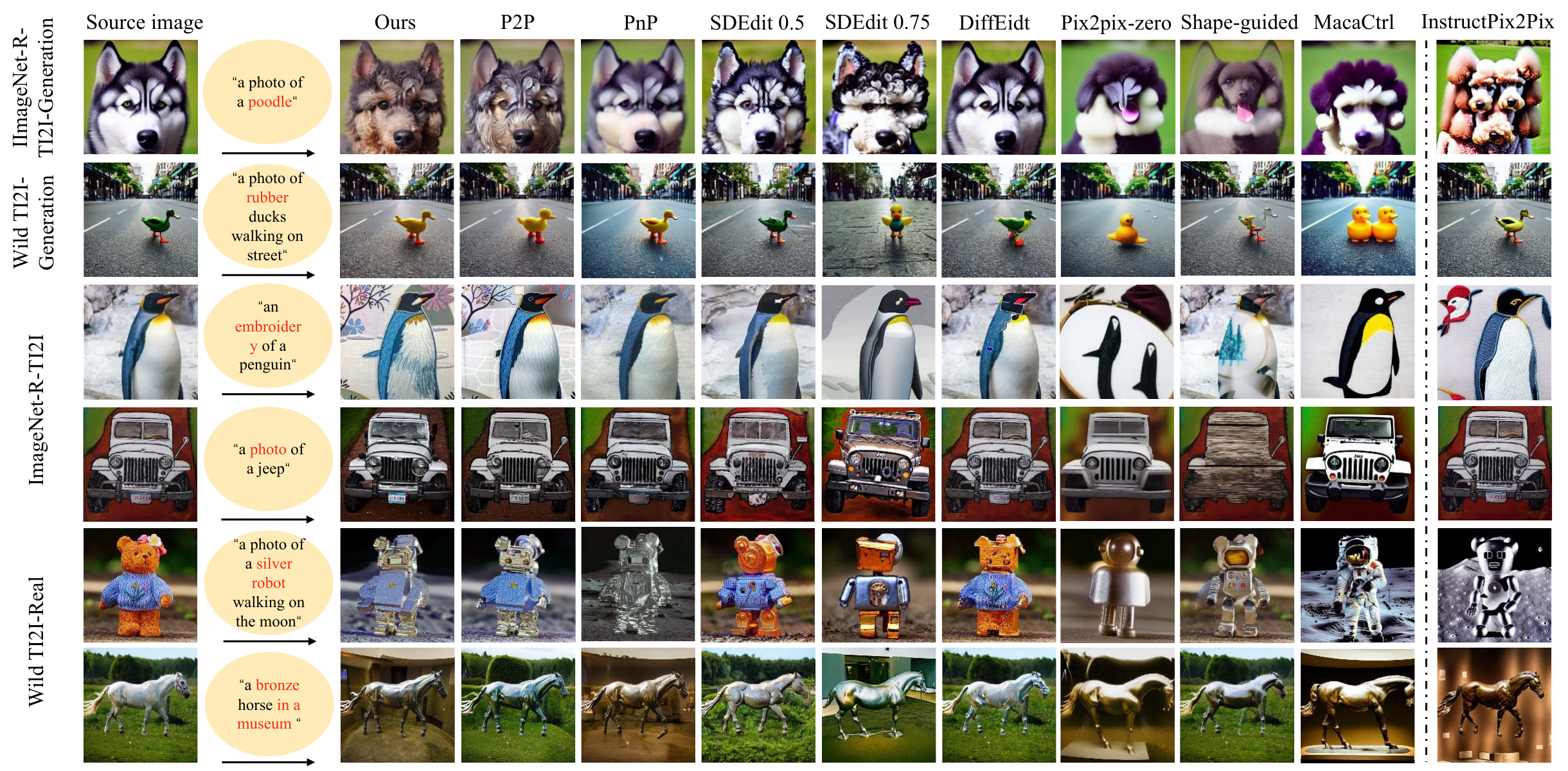}
\caption{Comparison to prior works. Left to right: source image, target prompt, our result, P2P~\cite{p2p}, PnP~\cite{pnp}, SDEdit~\cite{SDEdit} w/ two noising levels, DiffEdit~\cite{diffedit}, Pix2pixzero~\cite{pix2pixzero}, Shape-guided~\cite{Shape-guided}, MasaCtrl~\cite{masactrl} and InstructPix2Pix~\cite{instructpix2pix} (fine-tuning based method.)}
\label{fig:Comparison_results}
\end{figure*}

\begin{table*}[htp]
\centering
\aboverulesep=0pt
\belowrulesep=0pt
\scriptsize
\resizebox{0.84\textwidth}{!}{
\begin{tabular}{l|cc|cc||cc|cc||cc}
\toprule
\multirow{2}{*}{\textbf{Method}} & \multicolumn{2}{c|}{\textbf{ImageNet-R-TI2I fake}}&\multicolumn{2}{c||}{\textbf{ImageNet-R-TI2I real}}&\multicolumn{2}{c|}{\textbf{Wild-fake}}&\multicolumn{2}{c||}{\textbf{Wild-real}}&\multicolumn{2}{c}{\textbf{Editing Times $(s)$}}  \\

\cline{2-11}& \textbf{CS $\uparrow$ } & \textbf{CDS $\uparrow$  }  & \textbf{CS $\uparrow$ } & \textbf{CDS $\uparrow$  }  & \textbf{CS $\uparrow$  } & \textbf{CDS $\uparrow$  } & \textbf{CS $\uparrow$  } & \textbf{CDS $\uparrow$ }  & \textbf{fake $\downarrow$ } & \textbf{real $\downarrow$ } \\
\hline
SDEdit (0.5)	&- &-	&28.37	&0.1415	   &-   &-	    &27.48   &0.1220	& - &  \textbf{2.59}  \\     
SDEdit (0.75)	&- &-	&\underline{30.17}	&\underline{0.2171}	    &-   &-	&\textbf{29.79}   &0.2007	&- & \underline{3.35}  \\  
Shape-Guided &-& - &26.01 & 0.1090  &-& - &26.53& 0.1330 & - & 16.02 \\ 
DiffEdit &26.68  &0.0748 &26.50	&0.0909  &25.59   &0.0794	&26.33   &0.0879	 &9.02 & 4.85  \\ 
Pix2pixzero &27.94 & 0.2271 &28.96 & 0.1415 &28.19 &0.2864 &\underline{29.55} & 0.1462 & 24.92& 36.76\\ 
P2P &28.88	&\underline{0.3394}   &28.56	&0.2146	   &27.85    &0.2796	&28.42   &0.1930		 &6.41 &  55.32 \\ 
PnP &28.83	&0.2318		&28.76	&0.2073	   &\underline{28.20}   &0.2838		&28.46   &0.2020	& 335.65& 384.26 \\ 
MasaCtrl	&\underline{29.66}  &0.3024	 	&\textbf{31.40}	&0.2170	  &\textbf{29.96}   &\textbf{0.3474} &29.33 &0.2101 & \textbf{6.18} & 10.90 \\ 
\hline
Ours &\textbf{29.79}	& \textbf{0.3559}	   	&29.05	&\textbf{0.2271}   &27.88    &\underline{0.3116}		&29.04     & \textbf{0.2234}	&\underline{6.30} &  10.75  \\ 

\bottomrule
\end{tabular}
}
\caption{Quantitative experimental results over Wild-TI2I and ImageNet-R-TI2I benchmarks, including real and generated guidance images. CS: Clip score~\cite{clip} and CDS: Clip Directional Similarity~\cite{clip,StyleGAN-NADA}. Editing Times: per-image/second}
\label{Comparison_results}
\end{table*}

\noindent\textbf{Comparison to Other Methods}
Further, we compare our method with other state-of-the-art (SOTA) image editing methods~\cite{p2p,pnp,SDEdit,diffedit,instructpix2pix,Shape-guided,pix2pixzero,masactrl} over the Wild-TI2I and ImageNet-R-TI2I benchmarks. 
The experimental results are presented in Figure~\ref{fig:Comparison_results}
and Table~\ref{Comparison_results}.
As shown in Figure~\ref{fig:Comparison_results}, our method successfully converts different inputs for both real and synthetic images. In all examples, our method achieves high-fidelity editing that aligns with the target prompt while preserving the original image's structural information to the greatest extent possible. In contrast, SDEdit and InstructPix2Pix struggle to preserve the structural information of the original image. SDEdit aligns the editing results better with the target prompt when there is high-level noise but fails in the presence of low-level noise. InstructPix2Pix retains consistency with the target prompt but loses the original structural information. DiffEdit and Pix2pix-zero also struggle to perform better editing based on the target prompt. 
Similarly, PnP achieves good editing results, but it is a two-step method that leads to significant computational overhead; editing a single image in a generated image editing scenario takes approximately 335.65 seconds. In contrast, our method only requires around 6.30 seconds on an A100 GPU with 40GB memory, as Table~\ref{Comparison_results} indicates.

Table~\ref{Comparison_results} presents the quantitative experimental results of different editing algorithms on the Wild-TI2I and ImageNet-R-TI2I benchmarks. From Table~\ref{Comparison_results}, it is evident that our method outperforms all others in terms of the CDS metric. This indicates that our method excels in preserving the spatial structure of the original image and performing editing according to the requirements of the target prompt, yielding superior results. Meanwhile, our method achieves a good balance between time consumption and effectiveness, as demonstrated in Table~\ref{Comparison_results}.

\begin{figure}[H]
\centering
\includegraphics[width=0.465\textwidth]{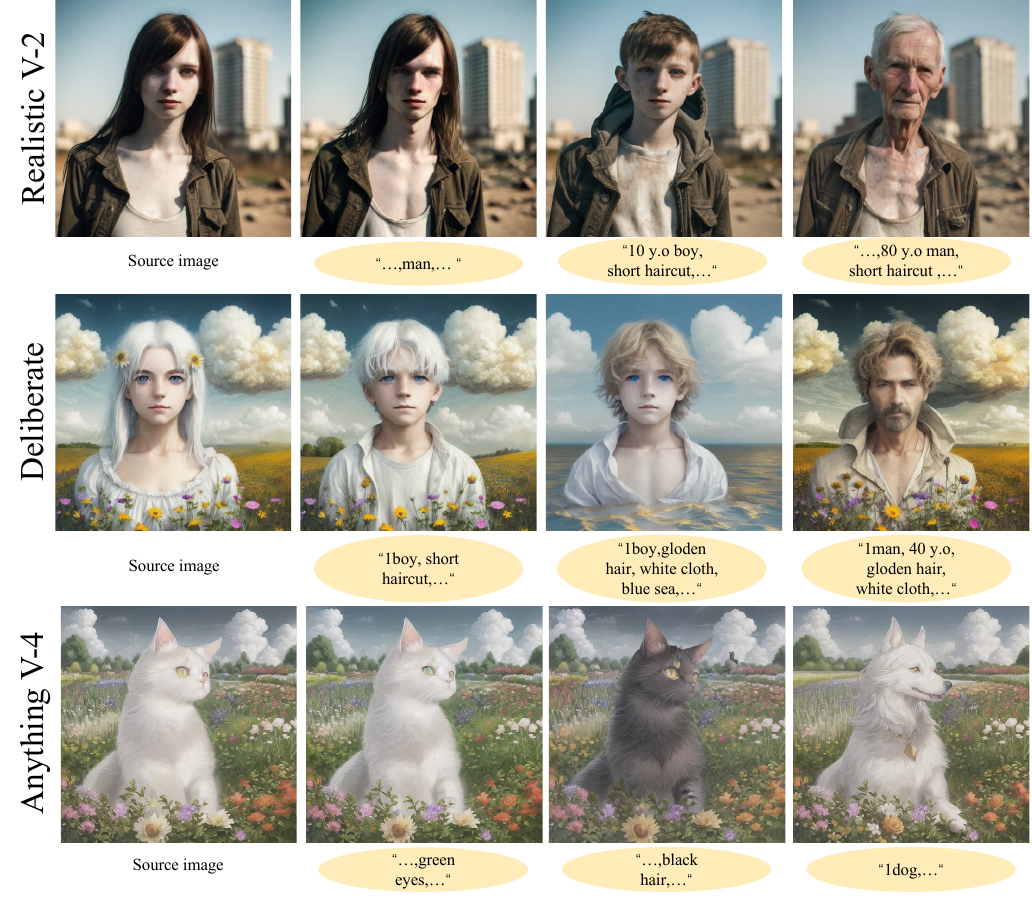}
\caption{Experimental results of our method using other TIS models, including Realistic-V2, Deliberate and Anything-V4.}
\label{fig:in_other_sd}
\end{figure}

\subsubsection{Results in Other TIS Models}
We have applied our method to other TIS models based on Stable Diffusion-style frameworks to demonstrate its transferability. Figure~\ref{fig:in_other_sd} showcases the editing results of our method on Realistic-V2, Deliberate, and Anything-V4 TIS models. From these results, it can be observed that our method is capable of effectively editing images on other diffusion models as well. For example, it can transform a girl into a boy, change a boy's age to 10 or 80, modify hairstyles, change hair colors, alter backgrounds, and switch categories.
\subsection{Limitations and Discussion}
Although our method employs probe analysis to elucidate the role of attention layers in the TIS model and proposes a novel method for editing images in multiple scenarios without complex operations, it still has some limitations. Firstly, our method is constrained by the generative capabilities of the TIS model. Our editing method will fail if the generative model cannot produce images consistent with the target prompt description. When editing real images, the original image must first be reconstructed. Some detailed information, especially facial details, may be lost during the reconstruction process, primarily due to the limitations of the VQ autoencoder~\cite{VAE}. Optimizing the VQ autoencoder is beyond the scope of this paper, as our objective is to provide a simple and universal editing framework. Addressing these challenges will be part of our future work.

\section{Conclusion}
In this work, we utilized probe analysis and conducted experiments to elucidate the following insights on TIS models: the cross-attention map carries the semantic information of the prompt, which leads to the ineffectiveness of image editing methods that rely on it. On the contrary, the self-attention map captures the spatial structural information of the original image, playing an essential role in preserving the image's inherent structure during editing. Based on our comprehensive analysis and empirical evidence, we have streamlined current image editing algorithms and proposed an innovative image editing approach. Our approach does not require additional tuning or the alignment of target and source prompts to achieve effective object or background editing in images. In extensive experiments across multiple datasets, our simplified method has outperformed existing image editing algorithms. Furthermore, our algorithm can be seamlessly adapted to other TIS models.

\paragraph{Acknowledgements}
This work is partially supported by Alibaba Cloud through the Research Talent Program with South China University of Technology, and the Program for Guangdong Introducing Innovative and Entrepreneurial Teams (No. 2017ZT07X183).

\newpage
{
    \small
    \bibliographystyle{ieeenat_fullname}
    \bibliography{main}
}

\clearpage
\setcounter{page}{1}
\maketitlesupplementary
The Supplementary Material of this paper consists of several sections that provide additional information and support for the main content. 
\begin{itemize}
    \item \textbf{Details of Data Collection:} Describes the data collection process used in the probing analysis and image editing experiments.
    \item \textbf{Probing Analysis Results:} Presents the complete results of the probing experiments, including supplementary experiments.
    \item \textbf{Impact of Replace Steps:} Provides ablation experimental results on replacing the self-attention map and cross-attention map at different denoising steps.
    \item \textbf{Real Image Editing with Null-text Inversion:} Introduces the method of using null-text inversion for editing real images and displays some experimental results.
\end{itemize}

\section{Details of Data Collection}
\label{sup:datasets}

\subsection{Data Collection for Probing Analysis}
\label{subsec:data_collection_probing_analysis}

\noindent\textbf{Cross-Attention Map Data:}
We have constructed five datasets by saving the cross-attention maps of the target word in the prompt, each containing 2,000 samples, for cross-attention map analysis. The prompts include color adjectives and animal nouns. Specifically, for color adjectives, we used two types of prompts: \emph{``a \textless{}color\textgreater{} car''} and \emph{``a \textless{}color\textgreater{} \textless{}object\textgreater{}''}, to construct the data. The prompt \emph{``a \textless{}color\textgreater{} car''} consists of ten categories of color words using 200 random seeds. We obtained the cross-attention maps by averaging the steps during the generation process. Each cross-attention map consists of 16 attention maps corresponding to the 16 attention layers of the diffusion model. The same procedure was followed for the construction of cross-attention maps in the format \emph{``a \textless{}color\textgreater{} \textless{}object\textgreater{}''}, but with two random seeds and 100 everyday objects. Similarly, we used the prompt format \emph{``a \textless{}animal\textgreater{} standing in the park''}, sampled 200 random seeds, and constructed 2,000 samples for animal nouns. For the cross-attention maps corresponding to non-editing words, we used the prompt format \emph{``a \textless{}color\textgreater{} car''} and saved the cross-attention maps for the words ``a'' and ``car.'' The same approach was applied to the complex text templates.

\noindent\textbf{Self-Attention Map Data:}
For the self-attention maps, we constructed two datasets, each containing 2,000 samples, using prompt formats \emph{``a \textless{}color\textgreater{} car''} and \emph{``a \textless{}animal\textgreater{} standing in the park''} and sampling 200 random seeds. However, due to the large size of the self-attention maps, which are $4096 \times 4096$, $1024 \times 1024$, $256 \times 256$, $64 \times 64$, and $8 \times 8$ in dimensions, we resized the layers with dimensions larger than or equal to $256 \times 256$ to $256 \times 256$ for the probing analysis experiments.

The 10 color, 10 animal, 100 object categories, and 12 more complex text templates are as follows:
\begin{lstlisting}[basicstyle=\scriptsize\ttfamily]
color_words = [
    "red", "blue", "green", "yellow", "brown", 
    "pink", "purple", "black", "white", "orange"
]
animal_list = [ 
    "dog", "giraffe", "horse", "lion", "rabbit", 
    "sheep", "cat", "monkey", "leopard", "tiger"
]
object_list = [ 
    "apple", "banana", "carrot", "dog", "flower", 
    "giraffe", "hat", "car", "train", "bicycle",
    ...
    "yak", "acorn", "bear", "caterpillar", "turtle", 
    "dandelion", "elephant", "feather", "grape", "hedgehog",
    "inchworm", "jackfruit", "kiwi", "lemon", "zebra",
    "mushroom", "otter", "peacock", "rose", "strawberry",
    "volcano", "watermelon", "xenops", "yucca", "cup"
]
complex_templates = [ 
    "a photo of a {} car and a dog",
    "a photo of {} car",
    "the painting of a {} car",
    ...
    "a cool painting of an old {} car",
    "a man and a {} car",
    "a {} car and a dog",
]
\end{lstlisting}

\subsection{Data Collection for Editing Experiments}
\label{sub:data_collection_editing_experiments}

\noindent\textbf{Car-fake-edit:}
Using the prompt format \emph{``a \textless{}color\textgreater{} car''} and 28 color words, we constructed Car-fake-edit, which contains 756 prompt pairs.

\noindent\textbf{Car-real-edit:}
We sampled 123 real images from the Stanford Car dataset~\cite{stanfordcar} with image sizes ranging from 512 to 768. We then used CLIP~\cite{clip} to align these images with the 28 color words, resulting in the source prompts for the original images in the format \emph{``a \textless{}color\textgreater{} car''}. We constructed 27 target prompts for each image, resulting in 3,321 image-text pairs.

\noindent\textbf{ImageNet-fake-edit:}
The paper FlexIT~\cite{LPIPS} proposes constructing a validation set from the ImageNet~\cite{imagenet_date} validation set. The method for constructing the test set is as follows: A subset of 273 labeled categories is taken from ImageNet, and these categories are manually divided into 47 clusters. During testing, only transformations within the same cluster are allowed; for example, a cat to a dog, but not a laptop to a butterfly. For each label T, eight random categories are sampled from the cluster to serve as queries, resulting in 2,184 queries for the 273 categories. We utilized their test dataset, which consists of 1,092 queries, and added ten animal categories to construct 1,182 prompt pairs using prompt templates.

\noindent\textbf{ImageNet-real-edit:}
Based on ImageNet-fake-edit, we used the prompt format ``a photo of a/an \{\}'' and constructed 1,092 image-text pairs for the real images. The color words, animal list, and prompt templates are as follows:
\begin{lstlisting}[basicstyle=\scriptsize\ttfamily]
color_words = [
    "red", "green", "blue", "yellow", 
    "orange", "purple", "pink", "black", 
    "white", "gray", "brown", "beige", 
    "cyan", "magenta", "teal", "lime",
    "olive", "navy", "maroon", "silver",
    "gold", "bronze", "peach", "coral",
    "indigo", "violet", "turquoise", "chocolate"
]
animal_list = [ 
    "dog", "giraffe", "horse", "lion", "rabbit", 
    "sheep", "cat", "monkey", "leopard", "tiger"
]
imagenet_templates = [
    "a photo of a {}",
    "a rendering of a {}",
    "a cropped photo of the {}",
    "the photo of a {}",
    "a photo of a clean {}",
    "a photo of a dirty {}",
    "a dark photo of the {}",
    "a photo of my {}",
    "a photo of the cool {}",
    "a close-up photo of a {}",
    "a bright photo of the {}",
    "a cropped photo of a {}",
    "a photo of the {}",
    "a good photo of the {}",
    "a photo of one {}",
    "a close-up photo of the {}",
    "a rendition of the {}",
    "a photo of the clean {}",
    "a rendition of a {}",
    "a photo of a nice {}",
    "a good photo of a {}",
    "a photo of the nice {}",
    "a photo of the small {}",
    "a photo of the weird {}",
    "a photo of the large {}",
    "a photo of a cool {}",
    "a photo of a small {}",
    "a {} in the park"
]
\end{lstlisting}

\section{Probing Analysis Results}
In this section, we present the complete experimental results for probe analysis and supplementary experiments, shown in Tables~\ref{sup:cross_attn_result}, \ref{sup:cross_attn_result_other_token}, \ref{sup:self_attn_result} and \ref{sup:cross_attn_result_more_complex}. 

Table~\ref{sup:cross_attn_result} presents the probe analysis experiment results for cross-attention maps with four sub-tables. The first three sub-tables, from top to bottom, correspond to the prompt formats \emph{``a/an \textless{}animal\textgreater{} standing in the park,''} \emph{``a \textless{}color\textgreater{}\textless{}object\textgreater{},''} and \emph{``a \textless{}color\textgreater{} car.''} The bottom sub-table corresponds to the experiment with training data \emph{``a \textless{}color\textgreater{} car''} and test data \emph{``a \textless{}color\textgreater{}\textless{}object\textgreater{}.''} 

For the experiments within the distribution, regardless of the prompt format, the cross-attention maps corresponding to the words can be accurately classified by the classifier. Even on out-of-distribution data, an average accuracy of around 50\% can be achieved. This indicates that the diffusion model's 16 layers of cross-attention maps can serve as good feature representations, containing semantic information about the corresponding words. This is consistent with the conclusion in the main text that the cross-attention maps are both weight matrices and rich in semantic information. Table~\ref{sup:cross_attn_result_other_token} presents the probe experiment results for non-target words, aiming to verify whether the attention maps corresponding to words other than the target word in the prompt contain the semantics of the target word. We conducted experiments using the simple prompt format \emph{``a \textless{}color\textgreater{} car.''}

Table~\ref{sup:self_attn_result} presents the complete probe experiment results for self-attention maps. Compared to cross-attention maps, self-attention maps are not directly usable as feature representations for classification, especially for prompts with color adjectives where the classification performance could be improved. However, compared to color adjective prompts, higher classification accuracy is observed for prompts with animal adjectives, which may be related to the self-attention maps' ability to represent objects' appearance contours in animal class images.

We expanded our investigation through probing analyses utilizing a set of twelve intricate text templates to eliminate the potential experimental bias that may arise from using consistently simple and regular templates in previous experiments. Examples of these templates include phrases such as \emph{``a painting of a wooden \textless{}color\textgreater{} car''} and \emph{``a photo of a \textless{}color\textgreater{} car and a dog''}, among others. The findings, as presented in Table~\ref{sup:cross_attn_result_more_complex}, corroborate the conclusions drawn from experiments using simpler text prompts, indicating consistency in results across varying levels of template complexity.

\label{sub:Probing Analysis}
\begin{table*}[h]
\centering
\aboverulesep=0pt
\belowrulesep=0pt
\begin{small}
\resizebox{\textwidth}{!}{
\begin{tabular}{lcccccccccccccccc|c}
\toprule
Class & Layer 1 & Layer 2 & Layer 3 & Layer 4 & Layer 5 & Layer 6 & Layer 7 & Layer 8 & Layer 9 & Layer 10 & Layer 11 & Layer 12 & Layer 13 & Layer 14 & Layer 15 & Layer 16 & Avg.\\
\hline
dog & 0.97 & 0.92 & 1.00 & 0.89 & 0.97 & 1.00 & 1.00 & 1.00 & 1.00 & 1.00 & 0.54 & 0.89 & 1.00 & 0.76 & 0.68 & 1.00 & 0.91 \\
giraffe & 0.94 & 0.97 & 1.00 & 0.97 & 1.00 & 1.00 & 1.00 & 1.00 & 0.92 & 0.97 & 0.97 & 1.00 & 0.97 & 0.92 & 0.92 & 0.89 & 0.97 \\
horse & 0.98 & 0.98 & 0.96 & 0.84 & 0.82 & 1.00 & 1.00 & 1.00 & 1.00 & 1.00 & 0.64 & 0.64 & 0.93 & 1.00 & 0.89 & 0.91 & 0.91 \\
lion & 0.98 & 1.00 & 1.00 & 1.00 & 1.00 & 1.00 & 0.93 & 0.93 & 0.95 & 0.98 & 0.66 & 0.90 & 0.98 & 0.93 & 0.98 & 0.98 & 0.95 \\
rabbit & 0.85 & 0.87 & 1.00 & 0.98 & 1.00 & 1.00 & 0.96 & 0.96 & 0.98 & 0.91 & 0.49 & 0.98 & 0.96 & 0.74 & 0.96 & 1.00 & 0.91 \\
sheep & 0.93 & 0.82 & 0.97 & 1.00 & 0.95 & 1.00 & 0.97 & 1.00 & 1.00 & 1.00 & 0.97 & 1.00 & 0.93 & 0.90 & 0.80 & 0.97 & 0.95 \\
cat & 0.95 & 0.95 & 0.90 & 0.95 & 0.88 & 1.00 & 1.00 & 0.95 & 0.95 & 0.82 & 0.85 & 0.78 & 0.97 & 0.75 & 0.70 & 0.78 & 0.89 \\
monkey & 1.00 & 1.00 & 1.00 & 1.00 & 0.95 & 1.00 & 0.98 & 1.00 & 1.00 & 1.00 & 0.67 & 1.00 & 1.00 & 1.00 & 0.95 & 1.00 & 0.97 \\
leopard & 1.00 & 0.89 & 0.97 & 1.00 & 0.97 & 1.00 & 0.97 & 0.95 & 1.00 & 1.00 & 0.76 & 0.97 & 0.97 & 0.79 & 0.97 & 0.87 & 0.94 \\
tiger & 1.00 & 1.00 & 1.00 & 1.00 & 1.00 & 1.00 & 1.00 & 0.94 & 0.97 & 1.00 & 0.91 & 0.88 & 1.00 & 1.00 & 1.00 & 0.97 & 0.98 \\
\midrule
green & 0.91 & 0.84 & 0.93 & 0.89 & 0.98 & 0.91 & 0.96 & 0.96 & 0.91 & 0.96 & 0.93 & 0.67 & 0.98 & 0.38 & 0.47 & 0.60 & 0.83 \\
white & 0.88 & 0.15 & 0.97 & 0.91 & 1.00 & 1.00 & 1.00 & 0.97 & 0.94 & 0.97 & 0.97 & 0.97 & 1.00 & 0.61 & 0.88 & 0.85 & 0.88 \\
purple & 0.98 & 0.93 & 1.00 & 0.98 & 1.00 & 0.98 & 0.63 & 0.98 & 0.84 & 0.93 & 0.86 & 0.95 & 1.00 & 0.51 & 0.60 & 0.44 & 0.85 \\
brown & 0.97 & 0.82 & 0.97 & 0.95 & 1.00 & 0.92 & 0.95 & 0.95 & 0.89 & 0.92 & 0.95 & 0.97 & 0.95 & 0.34 & 0.84 & 0.79 & 0.89 \\
blue & 0.92 & 0.65 & 0.86 & 0.92 & 1.00 & 1.00 & 1.00 & 1.00 & 0.95 & 0.92 & 0.86 & 0.95 & 0.97 & 0.19 & 0.62 & 0.57 & 0.84 \\
orange & 0.94 & 0.72 & 0.97 & 1.00 & 1.00 & 1.00 & 0.92 & 0.89 & 0.94 & 0.92 & 0.86 & 0.89 & 0.94 & 0.94 & 1.00 & 0.83 & 0.92 \\
black & 0.93 & 0.80 & 0.85 & 0.78 & 1.00 & 0.95 & 0.97 & 0.90 & 0.95 & 0.93 & 0.60 & 0.75 & 1.00 & 0.65 & 0.85 & 0.82 & 0.86 \\
pink & 0.93 & 0.54 & 0.98 & 0.93 & 1.00 & 1.00 & 0.95 & 0.95 & 1.00 & 1.00 & 1.00 & 0.51 & 1.00 & 0.46 & 0.51 & 0.88 & 0.85 \\
yellow & 0.96 & 0.51 & 0.96 & 0.89 & 0.96 & 0.77 & 0.98 & 0.96 & 1.00 & 0.98 & 0.91 & 1.00 & 0.98 & 0.36 & 0.81 & 0.68 & 0.86 \\
red & 0.85 & 0.85 & 0.97 & 0.90 & 1.00 & 0.97 & 0.60 & 0.97 & 0.93 & 0.85 & 0.90 & 0.70 & 1.00 & 0.23 & 0.80 & 0.65 & 0.82 \\
\midrule
green & 1.00 & 1.00 & 0.97 & 0.97 & 1.00 & 0.97 & 0.47 & 0.68 & 0.80 & 0.93 & 1.00 & 0.97 & 0.97 & 0.50 & 0.75 & 0.95 & 0.87 \\
white & 1.00 & 0.89 & 1.00 & 0.91 & 1.00 & 0.82 & 0.18 & 0.64 & 0.44 & 0.84 & 0.60 & 0.73 & 0.98 & 0.56 & 0.69 & 0.89 & 0.76 \\
purple & 1.00 & 1.00 & 1.00 & 1.00 & 1.00 & 1.00 & 0.77 & 0.69 & 0.90 & 0.87 & 0.05 & 0.97 & 0.97 & 0.67 & 0.90 & 1.00 & 0.86 \\
brown & 1.00 & 1.00 & 0.97 & 1.00 & 0.97 & 0.78 & 0.62 & 0.34 & 0.09 & 0.66 & 0.97 & 0.97 & 0.94 & 0.84 & 0.75 & 1.00 & 0.81 \\
blue & 1.00 & 0.97 & 0.97 & 0.94 & 1.00 & 1.00 & 0.94 & 0.89 & 0.72 & 1.00 & 0.42 & 0.97 & 1.00 & 0.86 & 0.94 & 0.94 & 0.91 \\
orange & 1.00 & 1.00 & 0.95 & 1.00 & 1.00 & 0.97 & 0.84 & 0.97 & 0.82 & 0.92 & 0.89 & 1.00 & 1.00 & 1.00 & 1.00 & 1.00 & 0.96 \\
black & 1.00 & 0.95 & 0.97 & 0.85 & 1.00 & 1.00 & 0.97 & 0.77 & 0.79 & 0.95 & 0.08 & 0.95 & 1.00 & 0.85 & 0.97 & 0.85 & 0.87 \\
pink & 1.00 & 0.95 & 0.95 & 0.95 & 1.00 & 1.00 & 0.95 & 0.88 & 0.76 & 0.90 & 1.00 & 0.95 & 1.00 & 0.71 & 0.90 & 0.95 & 0.93 \\
yellow & 1.00 & 1.00 & 1.00 & 1.00 & 1.00 & 1.00 & 0.74 & 0.88 & 0.85 & 0.97 & 0.74 & 0.94 & 1.00 & 0.76 & 0.82 & 0.74 & 0.90 \\
red & 1.00 & 1.00 & 1.00 & 1.00 & 1.00 & 0.95 & 0.15 & 0.87 & 0.87 & 0.74 & 0.44 & 0.95 & 1.00 & 0.95 & 0.90 & 0.79 & 0.85 \\
\hline
\hline
green & 0.82 & 0.58 & 0.84 & 0.13 & 0.84 & 0.16 & 0.55 & 0.03 & 0.16 & 0.08 & 0.92 & 0.42 & 0.34 & 0.29 & 0.29 & 0.84 & 0.46 \\
white & 0.74 & 0.11 & 0.42 & 0.82 & 0.16 & 0.18 & 0.39 & 0.42 & 0.68 & 0.63 & 0.58 & 1.00 & 0.63 & 0.32 & 0.08 & 0.89 & 0.50 \\
purple & 0.53 & 0.11 & 0.79 & 0.47 & 0.84 & 1.00 & 0.26 & 0.08 & 0.08 & 0.24 & 0.00 & 0.92 & 0.95 & 0.08 & 0.13 & 0.29 & 0.42 \\
brown & 0.97 & 0.68 & 0.74 & 0.66 & 0.84 & 0.18 & 0.13 & 0.08 & 0.00 & 0.03 & 0.92 & 0.76 & 0.68 & 0.45 & 0.61 & 0.18 & 0.50 \\
blue & 0.87 & 0.68 & 0.11 & 0.13 & 1.00 & 0.61 & 0.05 & 0.32 & 0.16 & 0.18 & 0.05 & 0.34 & 0.87 & 0.03 & 0.47 & 0.34 & 0.39 \\
orange & 0.87 & 0.24 & 0.89 & 0.95 & 0.50 & 0.55 & 0.45 & 0.13 & 0.18 & 0.34 & 0.08 & 0.92 & 0.76 & 0.26 & 0.92 & 0.34 & 0.52 \\
black & 0.50 & 0.82 & 0.68 & 0.26 & 0.55 & 0.82 & 0.84 & 0.00 & 0.42 & 0.68 & 0.00 & 0.05 & 0.95 & 0.45 & 0.61 & 0.79 & 0.53 \\
pink & 0.58 & 0.16 & 0.32 & 0.66 & 0.76 & 0.32 & 0.34 & 0.82 & 0.61 & 0.11 & 1.00 & 0.11 & 0.95 & 0.18 & 0.61 & 0.97 & 0.53 \\
yellow & 0.16 & 0.08 & 0.66 & 0.68 & 0.55 & 0.05 & 0.55 & 0.26 & 0.63 & 0.71 & 0.76 & 0.79 & 0.79 & 0.42 & 0.16 & 0.39 & 0.48 \\
red & 0.74 & 0.05 & 0.84 & 0.74 & 0.82 & 0.61 & 0.00 & 0.32 & 0.84 & 0.42 & 0.63 & 0.61 & 0.97 & 0.76 & 0.45 & 0.34 & 0.57 \\
\bottomrule
\end{tabular}
}
\end{small}
\caption{Complete experimental results of attribute mining using cross-attention maps in the diffusion model are shown. The table displays the probing accuracy results for various prompt formats: \emph{``a/an \textless{}animal\textgreater{} standing in the park,''} \emph{``a \textless{}color\textgreater{} car,''} \emph{``a \textless{}color\textgreater{}\textless{}object\textgreater{}''} and involves training on data in the format \emph{``a \textless{}color\textgreater{} car''} and testing on data in the format \emph{``a \textless{}color\textgreater{}\textless{}object\textgreater{}.''} The results are presented from top to bottom in the corresponding table.
}
\label{sup:cross_attn_result}
\end{table*}

\begin{table*}[h]
\centering
\aboverulesep=0pt
\belowrulesep=0pt
\begin{small}
\resizebox{\textwidth}{!}{
\begin{tabular}{lcccccccccccccccc|c}
\toprule
Class & Layer 1 & Layer 2 & Layer 3 & Layer 4 & Layer 5 & Layer 6 & Layer 7 & Layer 8 & Layer 9 & Layer 10 & Layer 11 & Layer 12 & Layer 13 & Layer 14 & Layer 15 & Layer 16 & Avg.\\
\hline
green & 0.35 & 0.80 & 0.03 & 0.00 & 0.53 & 0.00 & 0.00 & 0.00 & 0.00 & 0.00 & 0.00 & 0.00 & 0.00 & 0.00 & 0.05 & 0.00 & 0.11 \\
white & 0.03 & 0.00 & 0.20 & 0.47 & 0.00 & 0.00 & 0.00 & 0.00 & 0.70 & 0.00 & 0.00 & 0.72 & 0.00 & 0.30 & 0.07 & 0.05 & 0.16 \\
purple & 0.00 & 0.00 & 0.00 & 0.00 & 0.00 & 0.00 & 0.00 & 0.00 & 0.00 & 0.00 & 0.00 & 0.00 & 0.00 & 0.03 & 0.00 & 0.00 & 0.00 \\
brown & 0.82 & 0.00 & 0.45 & 0.00 & 0.00 & 0.00 & 0.00 & 0.00 & 0.00 & 0.00 & 0.00 & 0.00 & 0.00 & 0.75 & 0.15 & 0.97 & 0.20 \\
blue & 0.00 & 0.00 & 0.00 & 0.00 & 0.00 & 0.00 & 0.00 & 0.07 & 0.00 & 0.00 & 0.07 & 0.03 & 0.20 & 0.00 & 0.00 & 0.00 & 0.02 \\
orange & 0.00 & 0.00 & 0.00 & 0.00 & 0.00 & 0.00 & 0.00 & 0.00 & 0.00 & 0.00 & 0.00 & 0.00 & 0.00 & 0.00 & 0.00 & 0.00 & 0.00 \\
black & 0.03 & 0.03 & 0.07 & 0.82 & 0.00 & 1.00 & 1.00 & 0.00 & 0.00 & 1.00 & 0.00 & 0.60 & 0.85 & 0.20 & 0.88 & 0.05 & 0.41 \\
pink & 0.00 & 0.00 & 0.00 & 0.00 & 0.00 & 0.00 & 0.00 & 0.00 & 0.00 & 0.00 & 0.00 & 0.00 & 0.00 & 0.07 & 0.03 & 0.00 & 0.01 \\
yellow & 0.00 & 0.03 & 0.12 & 0.00 & 0.00 & 0.00 & 0.00 & 0.00 & 0.00 & 0.00 & 0.03 & 0.00 & 0.00 & 0.00 & 0.00 & 0.00 & 0.01 \\
red & 0.00 & 0.00 & 0.50 & 0.00 & 0.50 & 0.00 & 0.00 & 0.93 & 0.82 & 0.00 & 1.00 & 0.00 & 0.35 & 0.00 & 0.05 & 0.00 & 0.26 \\
\midrule
green & 0.67 & 0.00 & 0.67 & 0.00 & 0.17 & 0.67 & 0.33 & 0.33 & 0.00 & 0.00 & 0.17 & 0.00 & 0.17 & 0.50 & 0.17 & 0.00 & 0.24 \\
white & 0.83 & 0.00 & 0.33 & 0.00 & 0.00 & 1.00 & 0.50 & 0.08 & 0.83 & 0.58 & 0.08 & 0.00 & 0.92 & 0.00 & 0.83 & 1.00 & 0.44 \\
purple & 0.75 & 0.00 & 0.25 & 0.00 & 0.50 & 0.00 & 0.12 & 0.00 & 0.00 & 0.00 & 0.00 & 0.00 & 0.50 & 0.25 & 0.00 & 0.00 & 0.15 \\
brown & 0.40 & 1.00 & 0.40 & 0.80 & 0.00 & 0.00 & 0.20 & 0.70 & 0.20 & 0.00 & 0.00 & 0.90 & 0.60 & 0.00 & 0.70 & 0.00 & 0.37 \\
blue & 0.44 & 0.00 & 0.22 & 0.00 & 0.11 & 0.33 & 0.00 & 0.22 & 0.11 & 0.00 & 0.00 & 0.00 & 0.00 & 0.00 & 0.00 & 0.00 & 0.09 \\
orange & 0.40 & 0.40 & 0.60 & 1.00 & 0.80 & 1.00 & 1.00 & 0.80 & 0.80 & 1.00 & 1.00 & 1.00 & 1.00 & 0.40 & 0.80 & 0.80 & 0.80 \\
black & 1.00 & 0.00 & 0.83 & 0.17 & 0.67 & 0.33 & 0.67 & 1.00 & 0.50 & 0.50 & 0.00 & 0.00 & 0.17 & 0.00 & 0.00 & 0.83 & 0.42 \\
pink & 0.12 & 0.00 & 0.38 & 0.38 & 0.12 & 0.12 & 0.38 & 1.00 & 0.00 & 0.00 & 0.00 & 0.00 & 0.00 & 0.00 & 0.62 & 0.00 & 0.20 \\
yellow & 0.62 & 0.00 & 0.50 & 0.00 & 0.12 & 0.25 & 0.00 & 0.00 & 0.00 & 0.00 & 0.00 & 0.12 & 0.00 & 0.25 & 0.00 & 0.00 & 0.12 \\
red & 0.25 & 0.00 & 0.38 & 0.12 & 0.62 & 0.88 & 0.62 & 1.00 & 0.75 & 0.12 & 0.00 & 0.12 & 0.62 & 0.38 & 0.00 & 0.00 & 0.37 \\
\bottomrule
\end{tabular}
}
\end{small}
\caption{Complete experimental results of attribute mining using cross-attention map in non-edit tokens.}
\label{sup:cross_attn_result_other_token}
\end{table*}

\begin{table*}[h]
\centering
\aboverulesep=0pt
\belowrulesep=0pt
\begin{small}
\resizebox{\textwidth}{!}{
\begin{tabular}{lcccccccccccccccc|c}
\toprule
Class & Layer 1 & Layer 2 & Layer 3 & Layer 4 & Layer 5 & Layer 6 & Layer 7 & Layer 8 & Layer 9 & Layer 10 & Layer 11 & Layer 12 & Layer 13 & Layer 14 & Layer 15 & Layer 16 & Avg.\\
\toprule
dog & 0.47 & 0.38 & 0.53 & 0.23 & 0.55 & 0.60 & 0.50 & 0.53 & 0.78 & 0.60 & 0.57 & 0.53 & 0.57 & 0.47 & 0.38 & 0.38 & 0.50 \\
giraffe & 0.28 & 0.47 & 0.47 & 0.60 & 0.68 & 0.62 & 0.75 & 0.60 & 0.78 & 0.72 & 0.62 & 0.57 & 0.55 & 0.68 & 0.42 & 0.57 & 0.59 \\
horse & 0.12 & 0.20 & 0.50 & 0.38 & 0.70 & 0.70 & 0.40 & 0.80 & 0.82 & 0.65 & 0.70 & 0.68 & 0.55 & 0.53 & 0.30 & 0.28 & 0.52 \\
lion & 0.07 & 0.17 & 0.38 & 0.38 & 0.20 & 0.15 & 0.35 & 0.28 & 0.17 & 0.23 & 0.40 & 0.47 & 0.42 & 0.33 & 0.35 & 0.25 & 0.29 \\
rabbit & 0.30 & 0.30 & 0.25 & 0.33 & 0.23 & 0.20 & 0.25 & 0.17 & 0.20 & 0.23 & 0.23 & 0.20 & 0.30 & 0.42 & 0.28 & 0.28 & 0.26 \\
sheep & 0.17 & 0.23 & 0.53 & 0.55 & 0.33 & 0.45 & 0.07 & 0.38 & 0.25 & 0.45 & 0.55 & 0.62 & 0.42 & 0.53 & 0.38 & 0.25 & 0.38 \\
cat & 0.07 & 0.17 & 0.07 & 0.17 & 0.20 & 0.15 & 0.12 & 0.23 & 0.28 & 0.12 & 0.25 & 0.33 & 0.25 & 0.17 & 0.23 & 0.15 & 0.19 \\
monkey & 0.33 & 0.33 & 0.28 & 0.38 & 0.28 & 0.38 & 0.07 & 0.62 & 0.38 & 0.45 & 0.28 & 0.30 & 0.33 & 0.33 & 0.33 & 0.33 & 0.33 \\
leopard & 0.72 & 0.70 & 0.47 & 0.62 & 0.57 & 0.65 & 0.38 & 0.42 & 0.57 & 0.60 & 0.47 & 0.47 & 0.42 & 0.65 & 0.62 & 0.60 & 0.56 \\
tiger & 0.38 & 0.38 & 0.23 & 0.42 & 0.35 & 0.12 & 0.23 & 0.28 & 0.55 & 0.20 & 0.53 & 0.45 & 0.35 & 0.42 & 0.50 & 0.53 & 0.37 \\
\midrule
green & 0.00 & 0.12 & 0.00 & 0.00 & 0.00 & 0.00 & 0.00 & 0.03 & 0.05 & 0.00 & 0.12 & 0.05 & 0.00 & 0.00 & 0.00 & 0.12 & 0.03 \\
white & 0.00 & 0.33 & 0.00 & 0.00 & 0.53 & 0.05 & 0.45 & 0.68 & 0.30 & 0.55 & 0.07 & 0.03 & 0.00 & 0.15 & 0.00 & 0.25 & 0.21 \\
purple & 0.00 & 0.03 & 0.00 & 0.00 & 0.28 & 0.60 & 0.00 & 0.07 & 0.05 & 0.35 & 0.10 & 0.07 & 0.03 & 0.50 & 0.00 & 0.00 & 0.13 \\
brown & 0.00 & 0.10 & 0.00 & 0.00 & 0.00 & 0.00 & 0.05 & 0.03 & 0.05 & 0.00 & 0.17 & 0.17 & 0.05 & 0.03 & 0.00 & 0.10 & 0.05 \\
blue & 0.72 & 0.07 & 0.57 & 0.55 & 0.10 & 0.00 & 0.30 & 0.05 & 0.12 & 0.05 & 0.28 & 0.33 & 0.65 & 0.00 & 0.35 & 0.05 & 0.26 \\
orange & 0.00 & 0.00 & 0.00 & 0.00 & 0.00 & 0.00 & 0.03 & 0.03 & 0.00 & 0.00 & 0.00 & 0.00 & 0.00 & 0.00 & 0.00 & 0.00 & 0.00 \\
black & 0.00 & 0.35 & 0.03 & 0.03 & 0.07 & 0.00 & 0.33 & 0.05 & 0.15 & 0.07 & 0.17 & 0.20 & 0.10 & 0.25 & 0.10 & 0.28 & 0.14 \\
pink & 0.35 & 0.20 & 0.65 & 0.55 & 0.17 & 0.03 & 0.00 & 0.38 & 0.47 & 0.28 & 0.45 & 0.50 & 0.57 & 0.03 & 0.80 & 0.35 & 0.36 \\
yellow & 0.00 & 0.00 & 0.00 & 0.00 & 0.03 & 0.42 & 0.00 & 0.15 & 0.07 & 0.05 & 0.05 & 0.00 & 0.03 & 0.30 & 0.15 & 0.07 & 0.08 \\
red & 0.00 & 0.00 & 0.00 & 0.00 & 0.00 & 0.15 & 0.33 & 0.03 & 0.28 & 0.20 & 0.05 & 0.00 & 0.03 & 0.20 & 0.00 & 0.10 & 0.08 \\
\bottomrule
\end{tabular}
}
\end{small}
\caption{Complete experimental results of attribute mining using self-attention maps in the diffusion model.}
\label{sup:self_attn_result}
\end{table*}

\begin{table*}[h]
\centering
\aboverulesep=0pt
\belowrulesep=0pt
\begin{small}
\resizebox{\textwidth}{!}{
\begin{tabular}{lcccccccccccccccc|c}
\toprule
Class & Layer 1 & Layer 2 & Layer 3 & Layer 4 & Layer 5 & Layer 6 & Layer 7 & Layer 8 & Layer 9 & Layer 10 & Layer 11 & Layer 12 & Layer 13 & Layer 14 & Layer 15 & Layer 16 & Avg.\\
\toprule
green & 0.81 & 0.74 & 0.85 & 0.81 & 0.91 & 0.68 & 0.79 & 0.85 & 0.89 & 0.87 & 0.94 & 0.62 & 0.96 & 0.43 & 0.64 & 0.40 & 0.76 \\
white & 0.82 & 0.79 & 0.91 & 0.91 & 1.00 & 0.82 & 0.97 & 0.91 & 0.91 & 0.73 & 0.97 & 0.94 & 0.97 & 0.45 & 0.61 & 0.55 & 0.83 \\
purple & 0.98 & 0.65 & 1.00 & 0.86 & 0.98 & 0.95 & 0.79 & 0.93 & 0.95 & 0.93 & 0.91 & 0.79 & 0.98 & 0.47 & 0.35 & 0.72 & 0.83 \\
brown & 0.96 & 0.76 & 0.98 & 0.93 & 0.98 & 0.80 & 0.84 & 0.98 & 0.96 & 0.80 & 0.89 & 0.96 & 0.98 & 0.73 & 0.82 & 0.87 & 0.89 \\
blue & 0.89 & 0.62 & 0.95 & 0.59 & 0.86 & 0.95 & 0.70 & 1.00 & 0.97 & 0.62 & 0.73 & 0.89 & 1.00 & 0.16 & 0.59 & 0.68 & 0.76 \\
orange & 1.00 & 0.61 & 0.97 & 0.95 & 0.92 & 1.00 & 0.79 & 0.87 & 0.84 & 0.71 & 0.79 & 0.82 & 1.00 & 0.47 & 1.00 & 0.87 & 0.85 \\
black & 0.93 & 0.75 & 0.95 & 0.88 & 1.00 & 0.88 & 0.95 & 0.93 & 0.93 & 0.82 & 0.88 & 0.82 & 1.00 & 0.68 & 0.85 & 0.93 & 0.88 \\
pink & 0.90 & 0.02 & 0.88 & 0.98 & 0.93 & 0.95 & 0.88 & 0.85 & 0.93 & 0.98 & 0.95 & 0.76 & 0.98 & 0.27 & 0.66 & 0.66 & 0.79 \\
yellow & 0.89 & 0.56 & 0.89 & 0.78 & 1.00 & 0.75 & 0.92 & 0.94 & 0.86 & 0.69 & 0.81 & 0.86 & 0.92 & 0.33 & 0.58 & 0.47 & 0.77 \\
red & 0.93 & 0.00 & 0.90 & 0.75 & 0.93 & 0.95 & 0.88 & 0.90 & 0.97 & 0.80 & 0.85 & 0.75 & 1.00 & 0.60 & 0.53 & 0.70 & 0.78 \\
\midrule
green & 0.00 & 0.00 & 0.00 & 0.00 & 0.36 & 0.51 & 0.73 & 0.42 & 0.60 & 0.20 & 0.44 & 0.00 & 0.29 & 0.00 & 0.18 & 0.00 & 0.23 \\
white & 0.00 & 0.00 & 0.00 & 0.00 & 0.03 & 0.03 & 0.12 & 0.03 & 0.03 & 0.03 & 0.06 & 0.03 & 0.03 & 0.03 & 0.36 & 0.00 & 0.05 \\
purple & 0.00 & 0.00 & 0.00 & 0.02 & 0.00 & 0.00 & 0.00 & 0.00 & 0.00 & 0.02 & 0.02 & 0.16 & 0.07 & 0.00 & 0.19 & 0.00 & 0.03 \\
brown & 0.00 & 0.00 & 0.00 & 0.00 & 0.00 & 0.00 & 0.03 & 0.03 & 0.05 & 0.11 & 0.08 & 0.00 & 0.00 & 0.00 & 0.05 & 0.00 & 0.02 \\
blue & 0.00 & 0.00 & 0.00 & 0.00 & 0.00 & 0.00 & 0.00 & 0.00 & 0.00 & 0.00 & 0.00 & 0.00 & 0.00 & 0.00 & 0.00 & 0.00 & 0.00 \\
orange & 0.00 & 0.00 & 0.00 & 0.00 & 0.00 & 0.00 & 0.03 & 0.00 & 0.00 & 0.42 & 0.19 & 0.00 & 0.00 & 0.00 & 0.08 & 0.00 & 0.05 \\
black & 0.00 & 0.00 & 0.00 & 0.00 & 0.03 & 0.03 & 0.12 & 0.07 & 0.07 & 0.07 & 0.10 & 0.00 & 0.03 & 0.00 & 0.03 & 0.00 & 0.03 \\
pink & 1.00 & 1.00 & 1.00 & 1.00 & 0.00 & 0.00 & 0.09 & 0.00 & 0.00 & 0.02 & 0.02 & 0.85 & 0.00 & 0.91 & 0.04 & 1.00 & 0.43 \\
yellow & 0.00 & 0.00 & 0.00 & 0.00 & 0.17 & 0.22 & 0.02 & 0.22 & 0.20 & 0.39 & 0.00 & 0.00 & 0.17 & 0.00 & 0.17 & 0.00 & 0.10 \\
red & 0.00 & 0.00 & 0.00 & 0.00 & 0.68 & 0.55 & 0.00 & 0.40 & 0.17 & 0.00 & 0.17 & 0.00 & 0.65 & 0.00 & 0.10 & 0.00 & 0.17 \\
\bottomrule
\end{tabular}
}
\end{small}
\caption{Probing accuracy of attention maps in more complex text templates. Upper: cross-attention map, Lower: self-attention map.}
\label{sup:cross_attn_result_more_complex}
\end{table*}

\section{Impact of Replacement Steps}
\label{sub:impact_replace_steps} 

\begin{figure*}[htp]
\centering
\includegraphics[width=0.95\textwidth]{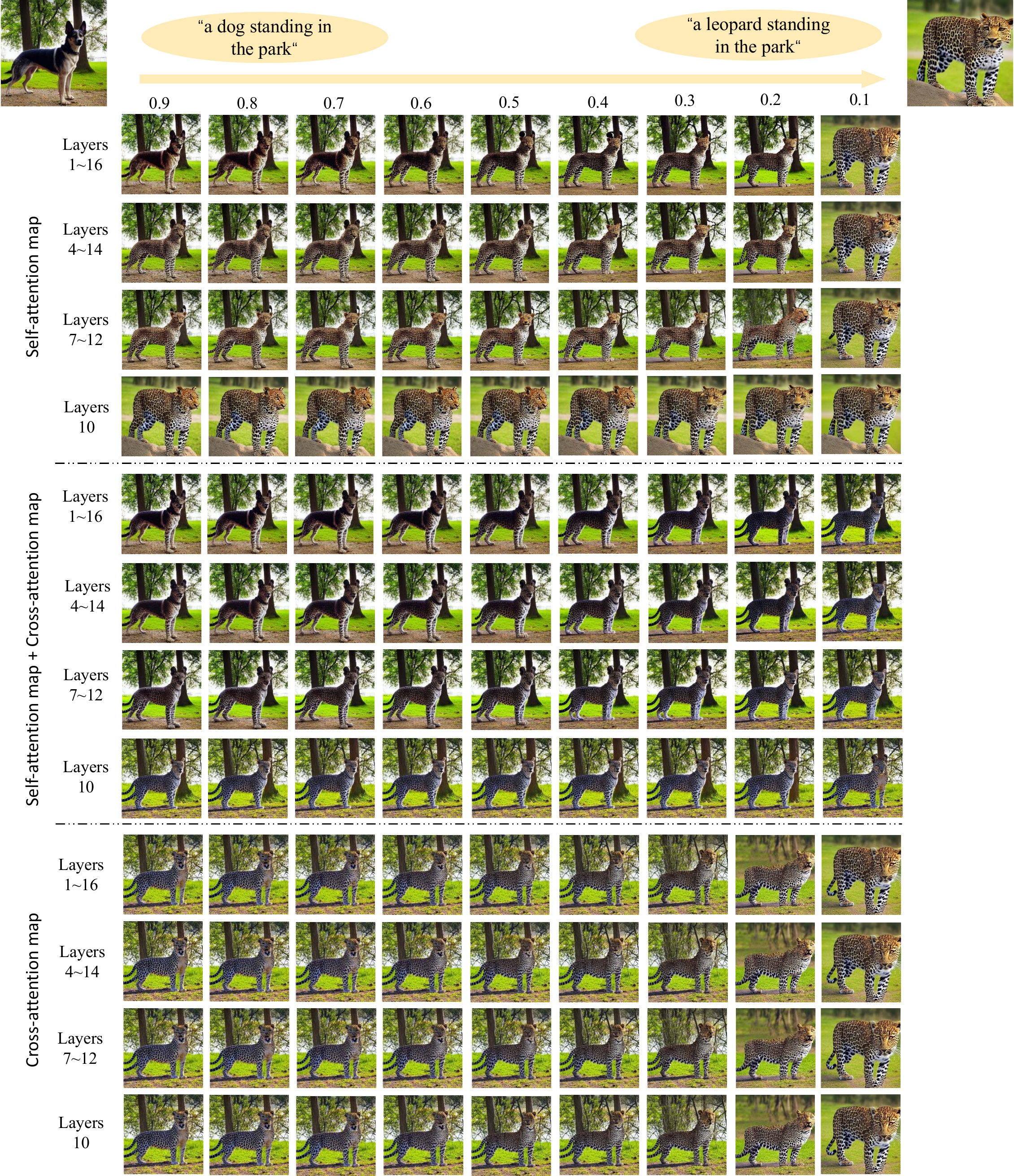}
\caption{Ablation experiment results showing three types of attention map replacements for noun editing at different replace step ratios. A higher ratio denotes a greater number of replacement steps.}
\label{sup:ablation_nouns_steps}
\end{figure*}

\begin{figure*}[htp]
\centering
\includegraphics[width=0.95\textwidth]{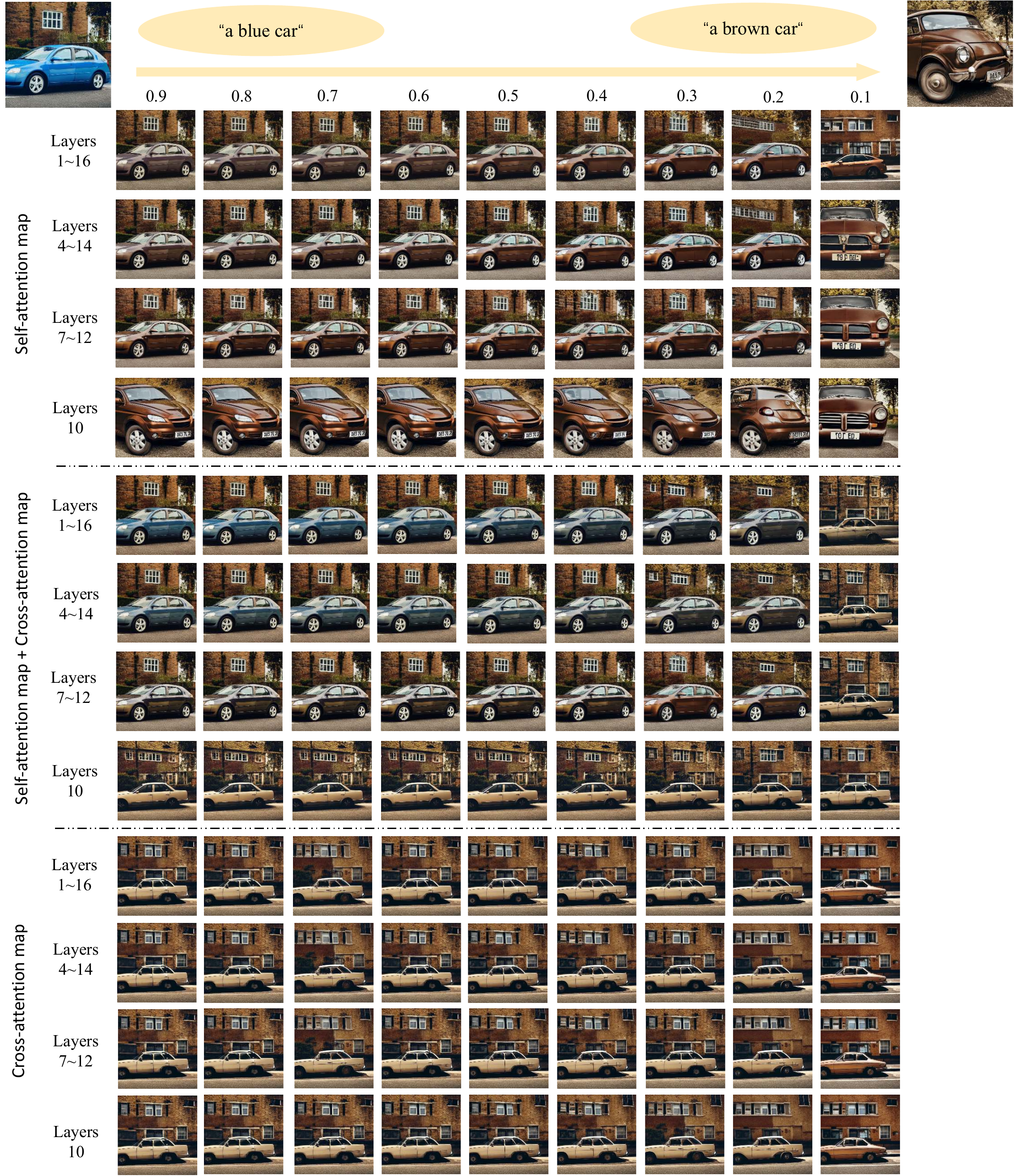}
\caption{Ablation experiment results showing three types of attention map replacements for color editing at different replace step ratios. A higher ratio denotes a greater number of replacement steps.}
\label{sup:ablation_color_steps}
\end{figure*}

In this section, we conduct ablation experiments on different attention layers of cross-attention and self-attention maps under various denoising steps. The experimental results are presented in Figure~\ref{sup:ablation_nouns_steps} and Figure~\ref{sup:ablation_color_steps}.

When only the cross-attention map is replaced during editing, the target image loses the structural information of the original image. As shown in Figure~\ref{sup:ablation_nouns_steps}, although the leopard in the target image bears a resemblance to the dog in the original image, notable modifications in the background, particularly the grass, are observed. Similarly, when performing a color conversion on a car using only the cross-attention map, the original image's structural information is lost, leading to a car that lacks its original structure and takes on a brown appearance.

When both the cross-attention map and the self-attention map are replaced simultaneously, the results depicted in Figures~\ref{sup:ablation_nouns_steps} and \ref{sup:ablation_color_steps} are obtained by keeping the cross-replace ratio fixed at 0.8 while varying the self-replace ratio. The replacement of the cross-attention map aids in swiftly identifying the target region and reconstructing the structure of the original image. However, it also introduces the original image's feature information, particularly when replacing attention maps in all layers, which significantly includes the original features. As illustrated in Figures~\ref{sup:ablation_nouns_steps} and \ref{sup:ablation_color_steps}, the leopard exhibits attributes similar to a dog, while the car retains its blue color.

When only the self-attention map is replaced with a low self-replace ratio, such as 0.1, the resulting target image closely resembles the one obtained using the target prompt directly. However, when the self-attention map is replaced in all attention layers and for 90\% of the denoising steps, a target image that closely matches the original image is generated, as depicted in the top left corner of Figure~\ref{sup:ablation_nouns_steps}. A more balanced approach involves replacing the self-attention map in layers 4–14 with replacement ratios ranging from 0.4 to 0.8, resulting in more favorable outcomes.

\section{Real Image Editing with Null-Text Inversion}
\label{sup:null_text_inversion} 

\begin{algorithm}[!ht]
    \renewcommand{\algorithmicrequire}{\textbf{Input:}}
    \renewcommand{\algorithmicensure}{\textbf{Output:}}
    \caption{Editing Method for Real Image Using Null-Text Inversion~\cite{null-text-inversion}}
    \label{sup:real_image_algorithm_null_text}
    \begin{algorithmic}[1] 
        \REQUIRE $P_{src}$: a source prompt;
                 $P_{dst}$: a target prompt;
                 $I$: real image;
                 $S$: random seed;
        \ENSURE  $I_{dst}$: edited image;
                 $I_{res}$: reconstructed image;
        \STATE $\hat{z}_{T},\{\varnothing\}_{t=1}^{T} \leftarrow \textsc{NullText-Inv}(P_{src},I)$;
        \STATE $z^{*}_{T} \leftarrow \hat{z}_{T}$;
        \FOR {$t = T,T-1,...,1$}
            \STATE $z_{t-1},M_{self} \leftarrow \textsc{DM}(\hat{z}_{T},\{\varnothing\}_{t},P_{src},t)$;
            \STATE $z^{*}_{t-1} \leftarrow \textsc{DM}(z^{*}_{t},\{\varnothing\}_{t},P_{dst},t)\{\!M^{*}_{self} \leftarrow M_{self}\}$;
        \ENDFOR
        \STATE \textbf{return} $(z_{0},z^{*}_{0})$.
    \end{algorithmic}
\end{algorithm}

\begin{figure*}[htp]
\centering
\includegraphics[width=0.95\textwidth]{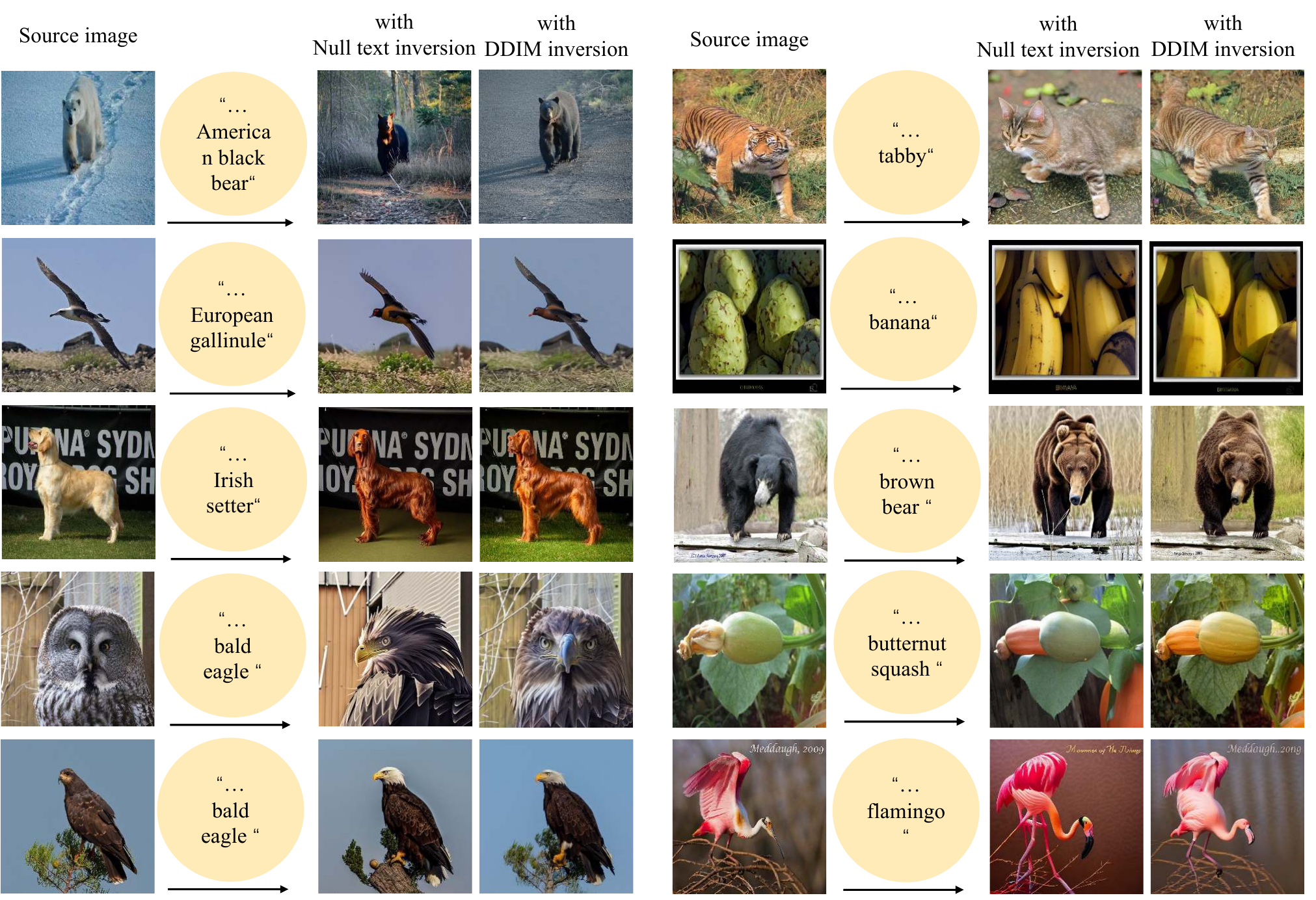}
\caption{Experiment results of real image editing in ImageNet-Real-Edit with Null-Text Inversion and DDIM Inversion. The source prompt form is \emph{``a photo of a/an \textless{}object\textgreater{}''} for Null-Text Inversion.}
\label{sup:ddim_nulltext_results}
\end{figure*}

Algorithm~\ref{sup:real_image_algorithm_null_text} outlines the pseudo-code for editing real images using the Freeprompt Editing (FPE) combined with Null-Text Inversion~\cite{null-text-inversion}. Figure~\ref{sup:ddim_nulltext_results} presents the experimental results of editing real images using DDIM Inversion~\cite{DDIM} and Null-Text Inversion. Both methods effectively modify the original image based on the target text.

\end{document}